\documentclass[letterpaper, 10 pt, conference]{ieeeconf}  %

\IEEEoverridecommandlockouts                              %

\usepackage{graphics} %
\usepackage{epsfig} %
\usepackage{times}
\usepackage{graphicx}
\usepackage{float}
\usepackage{times} 
\usepackage{amsmath} 
\usepackage{amssymb}  
\usepackage{booktabs}
\usepackage{epstopdf}
\usepackage{color}
\usepackage{url}
\usepackage{cite}
\usepackage{balance}
\usepackage{textcomp}
\usepackage{makecell}
\usepackage{threeparttable}

\usepackage{multirow}
\usepackage{caption}
\usepackage[table,xcdraw]{xcolor}
\usepackage{pifont}
\usepackage{hyperref}

\usepackage{soul}
\usepackage{tcolorbox}

\usepackage{pgf-pie}

\title{\LARGE \bf
DualMap: Online Open-Vocabulary Semantic Mapping for Natural Language Navigation in Dynamic Changing Scenes
}

\author{
  Jiajun Jiang$^{1}$, Yiming Zhu$^{1}$, Zirui Wu$^{1}$, Jie Song$^{1, 2}$
\thanks{$^{1}$The Hong Kong University of Science and Technology (Guangzhou)}
\thanks{$^{2}$The Hong Kong University of Science and Technology}
}

\begin{document}

\maketitle
\thispagestyle{empty}
\pagestyle{empty}

\begin{abstract}
We introduce DualMap, an online open-vocabulary mapping system that enables robots to understand and navigate dynamically changing environments through natural language queries. Designed for efficient semantic mapping and adaptability to changing environments, DualMap meets the essential requirements for real-world robot navigation applications.
Our proposed hybrid segmentation frontend and object-level status check eliminate the costly 3D object merging required by prior methods, enabling efficient online scene mapping. 
The dual-map representation combines a global \textit{abstract} map for high-level candidate selection with a local \textit{concrete} map for precise goal-reaching, effectively managing and updating dynamic changes in the environment.
Through extensive experiments in both simulation and real-world scenarios, we demonstrate state-of-the-art performance in 3D open-vocabulary segmentation, efficient scene mapping, and online language-guided navigation.
Project page: \href{https://eku127.github.io/DualMap/}{https://eku127.github.io/DualMap/}

\end{abstract}

\section{Introduction}
Imagine a robot navigating a bustling office when a colleague suddenly requests, \textit{``Bring me the MacBook I was using earlier.''} This seemingly simple task requires a mapping system that bridges human language and dynamic environments. To succeed, three critical capabilities emerge:
1)~\textbf{Open-vocabulary}: The system must detect objects in a class-agnostic manner during mapping and interpret various natural language queries (e.g., \textit{`the MacBook'}) without relying on a set of fixed categories;
2)~\textbf{Efficient online mapping}: The system must incrementally construct and maintain semantic maps in real-time, allowing it to adapt to ongoing changes in the environment;
3)~\textbf{Navigation with dynamic changes}: In human-centered environments, objects frequently change their location or state—for instance, a laptop might be moved or unfolded. The system should implement a navigation strategy that guides the agent to the relocated object and updates the map for lifelong navigation.

To support such tasks, existing works~\cite{hughes2022hydra, schmid2024khronos} construct semantic maps online from RGBD input using closed-set detectors. While efficient, these methods are constrained to predefined categories and cannot handle open-ended queries.
Recent efforts~\cite{Jatavallabhula2023ConceptFusion,gu2024conceptgraphs,werby23hovsg} address this limitation by integrating vision foundation models~\cite{Liu2023GroundingDM, Kirillov2023SegmentA, zhao2023fast}.
However, these methods assume static environments.
Other open-vocabulary systems~\cite{tang2025openin, yan2025dynamic} tackle dynamic changes but typically require significant time—often several hours—to construct a usable map, making them impractical for lifelong robotic navigation.
Consequently, no existing system simultaneously satisfies open-vocabulary understanding, efficient online mapping, and dynamic navigation, which restricts their applicability in real-world robotic applications.

\begin{figure}[t]
    \centering
    \includegraphics[width=1\linewidth]{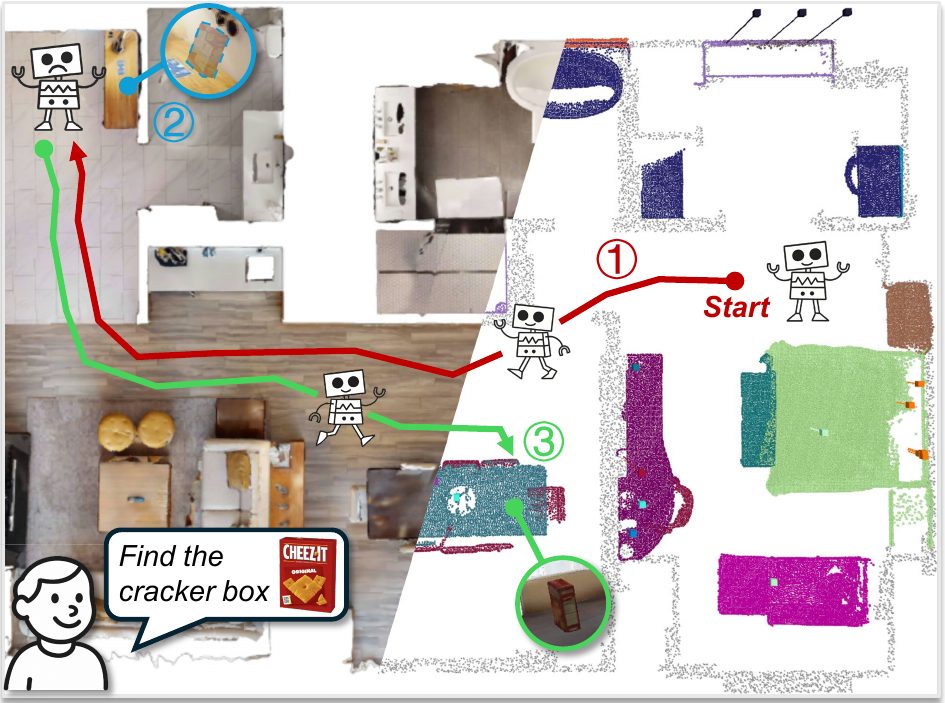}
    \caption{DualMap enables online open-vocabulary navigation under dynamic object changes. 
    Given a natural language query, the robot navigates toward the desk guided by the \textit{abstract map} (\ding{172}), only to discover that the queried object is missing (\ding{173}).
    The system updates the queried object's status and new observations (e.g., cracker box on the table) via the online-built \textit{concrete map} during navigation.
    This continuously updated map facilitates efficient re-selection of candidates to locate the relocated object (\ding{174}).}
    \label{fig:teaser}
    \vspace{-16pt}
\end{figure}

In this work, we present DualMap, an online open-vocabulary map representation that is the first to meet all three aforementioned requirements. To enable efficient open-vocabulary mapping, the system constructs a high-quality 3D semantic \textit{concrete map} through two key designs. A hybrid frontend process facilitates fast, holistic, and open-vocabulary object segmentation. Intra-object self-status checks, including stability checks and object splitting, help remove noisy segments and enhance map fidelity. These innovations also eliminate the need for the costly 3D inter-object merging commonly used in prior works~\cite{gu2024conceptgraphs, werby23hovsg}.

\begin{figure*}[t]
    \centering
    \includegraphics[width=1.0\textwidth]{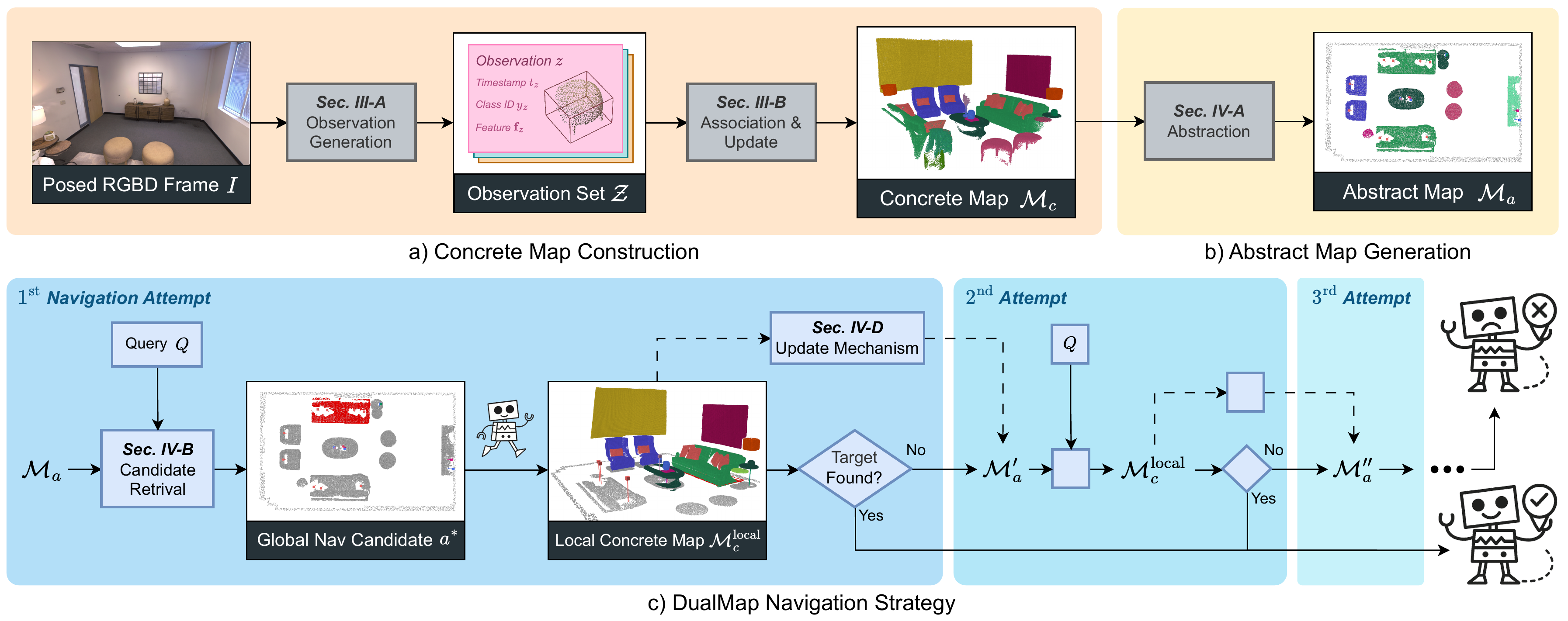}
    \caption{DualMap system overview: 
    a) A detailed 3D semantic concrete map $\mathcal{M}_c$ is built from online observations of posed RGBD frames;
    b) An anchor-based abstract map $\mathcal{M}_a$ is derived from $\mathcal{M}_c$, retaining global layout and static objects;
    c) Given a natural language query $Q$, the agent retrieves a global candidate $a^*$ from $\mathcal{M}_a$ and starts navigation. 
    During execution, it incrementally builds a local concrete map $\mathcal{M}_c^{\text{local}}$, checks for target object presence, and updates the abstract map $\mathcal{M}_a$ accordingly. 
    If the target is not found near the $a^*$, a new navigation attempt is made using the updated map $\mathcal{M}_a'$. This loop continues until the target is found or the attempt limit is reached.
}
    \label{fig:pipeline}
    \vspace{-14pt}
    
\end{figure*}

To support navigation in dynamic environments, we convert the fine-grained \textit{concrete map} into an \textit{abstract map} composed of typically static objects (e.g., beds, desks) and layout information. This abstraction is based on the insight that maintaining precise details of all objects globally is unnecessary for navigation: global structural cues are sufficient for high-level planning, while object details can be retrieved locally through online perception. We then propose a dual-map navigation strategy, where the agent uses the global abstract map for candidate selection and the local concrete map for accurate object localization. The abstract map is continuously updated during navigation, facilitating efficient re-selection when the queried object has moved.
Experimental results on Replica~\cite{straub2019replica}, ScanNet~\cite{dai2017scannet}, and HM3D~\cite{yadav2022habitat} demonstrate that DualMap achieves strong performance in semantic mapping, efficiency, and object navigation across both static and dynamic settings. Real-world experiments in indoor and outdoor scenes confirm its practical applicability. The code will be made available to support future research.

\section{Related Work}
\label{related}

\subsection{Open-Vocabulary Online Semantic Mapping}
Fast closed-set detectors like YOLO~\cite{redmon2016you} have been widely used in semantic robotic mapping~\cite{mccormac2017semanticfusion, mccormac2018fusion++, rosinol2020kimera, hughes2022hydra}, but their fixed category sets limit generalization to open-world settings.
To enable open-vocabulary perception, two main approaches have emerged.
One uses image tagging models~\cite{zhang2024recognize, huang2023tag2text} to generate class labels from incoming frames, which then guide closed-set detectors~\cite{cheng2024yolo} for object detection.
ConceptGraphs~\cite{gu2024conceptgraphs} follows this pipeline but remains offline due to the high cost of running GroundedSAM~\cite{ren2024grounded} per frame.
A more common alternative applies class-agnostic segmentation~\cite{Kirillov2023SegmentA,zhao2023fast} directly, followed by feature embedding using vision foundation models (VFMs)~\cite{radford2021clip, zhai2023sigmoid, mobileclip2024}. 
HOV-SG~\cite{werby23hovsg} uses this method and merges over-segmented objects at the 3D level, improving semantic accuracy but remaining expensive in computation.
CLIO~\cite{maggio2024clio} addresses efficiency by merging only task-relevant objects. However, since the map is task-specific, querying non-task objects requires extra map reconstruction, limiting map reusability.
Different from prior approaches, DualMap adopts a hybrid segmentation strategy for efficient, holistic open-vocabulary mapping and replaces costly 3D inter-object merging with lightweight intra-object checks, enabling online operation.

\begin{table}[t!]
\centering
\caption{Comparison of recent semantic mapping systems.}
\label{tab:semantic-comparison}
\resizebox{\columnwidth}{!}{%
\begin{tabular}{lccc}
\toprule
\textbf{System} & \textbf{Open-Vocab} & \textbf{Online Mapping} & \textbf{Dynamic Handling} \\
\midrule
Hydra~\cite{hughes2022hydra} &                    & \checkmark & \\
Khronos~\cite{schmid2024khronos} &                  & \checkmark & \checkmark \\
ConceptGraphs~\cite{gu2024conceptgraphs} & \checkmark &             & \\
HOV-SG~\cite{werby23hovsg} & \checkmark       &             & \\
CLIO~\cite{maggio2024clio}   & \checkmark       & \checkmark  & \\
OpenIN~\cite{tang2025openin} & \checkmark      &             & \checkmark \\
\rowcolor{gray!20}
DualMap (Ours) & \checkmark & \checkmark & \checkmark \\
\bottomrule
\end{tabular}
}

\vspace{-15pt}

\end{table}

\subsection{Mapping and Navigation in Dynamic Environments}

Previous 3D semantic mapping methods~\cite{zhang20233d, gu2024conceptgraphs, werby23hovsg} support closed-set or open-vocabulary object navigation, but assume static environments. While being effective in static settings, they struggle to handle dynamic object movements common in real-world scenarios.
To address this challenge, recent works~\cite{schmid2022panoptic, fu2022planesdf, schmid2024khronos, tang2025openin} have explored ways to adapt semantic maps to dynamic scene changes.
OpenIN~\cite{tang2025openin} proposes an open-vocabulary carrier-relation scene graph that updates memory when objects are relocated, but its offline pipeline limits update speed and hinders long-term operation.
Khronos~\cite{schmid2024khronos} introduces a real-time semantic mapping system for dynamic environments, maintaining a consistent dense mesh. However, it relies on closed-set segmentation and lacks support for arbitrary language queries, and also lacks navigation strategies for locating moved queried objects.
The proposed DualMap system addresses the dynamic navigation challenge with a dual-map design. The system updates memory online during operation, enabling efficient navigation candidate re-selection when the queried object moves, and supports lifelong language-guided navigation.

\section{Online Concrete Map}
\label{concrete_map}

In this section, we detail the construction of the online concrete map $\mathcal{M}_c$ (Fig.~\ref{fig:pipeline}-a). We first describe how to efficiently generate object-level observations from the RGBD stream using fast, holistic open-vocabulary segmentation. Then, we explain how these observations incrementally update the map to ensure consistent and accurate scene representation.

\subsection{Observation Generation}
\subsubsection{Hybrid Open-Vocabulary Segmentation} 

To efficiently detect and segment all potential objects in the scene, we adopt a hybrid open-vocabulary segmentation pipeline. 
As in Fig.~\ref{fig:obs}-a, we employ YOLO~\cite{cheng2024yolo} to rapidly produce object-level detections, and further generate the corresponding segmentation masks using MobileSAM~\cite{zhang2023faster}. Since YOLO requires a predefined category list, a task-specific detection list is generated once at the beginning of operation by prompting a large language model~\cite{achiam2023gpt} based on the robot's working context (e.g., \textit{``Give me a list of common objects when the robot operates in an office.''}). We further refine YOLO outputs by merging overlapping bounding boxes with similar color distributions, measured via histogram similarity in RGB space. 
In parallel (Fig.~\ref{fig:obs}-b), we run FastSAM~\cite{zhao2023fast}, an open-set segmentation model, to capture objects beyond YOLO's categories. We prioritize YOLO detections during fusion, retaining all YOLO outputs and supplementing them with non-overlapping FastSAM segments ($\oplus$ in Fig.~\ref{fig:obs}). This hybrid strategy ensures holistic and efficient object-level segmentation suitable for downstream mapping.

\subsubsection{Semantic Feature Embedding}
\label{sec:feat-weight}
We generate a semantic feature by combining visual and textual information. 
For each detected object, we crop the image region defined by its 2D bounding box and encode it using CLIP~\cite{radford2021clip} to obtain the image feature $\mathbf{f}_\text{image}$. If a class label is available (e.g., \textit{`chair'}), it is also encoded into a CLIP text feature $\mathbf{f}_\text{text}$. The final segment feature $\mathbf{f}$ is computed as a weighted sum:
\begin{equation}
\label{qua:feat-weight}
\mathbf{f} = w_{\text{image}} \mathbf{f}_{\text{image}} + w_{\text{text}}\mathbf{f}_{\text{text}}, \qquad \mathbf{f}_{\{\text{image, text}\}} \in \mathbb{R}^d
\end{equation}
where $w_{\text{image}} = 0.7$ and $w_{\text{text}} = 0.3$. For objects labeled as ``\texttt{null}'' by FastSAM, we replace $\mathbf{f}_{\text{text}}$ with the normalized mean text feature computed over all known class embeddings to eliminate semantic bias.

\subsubsection{Observation Structure}

After processing current frame $I$, we generate an observation set $\mathcal{Z}$, defined as $\mathcal{Z} = \{ z_1, z_2, \ldots, z_N \}$, where $N$ denotes the number of segments. Each observation $z = (\mathbf{P}_z, \mathbf{f}_z, y_z, t_z) \in \mathcal{Z}$ is structured with the following components:
The 3D point cloud $\mathbf{P}_z$ is generated by projecting the segmented region from the depth map into the world coordinate system using the camera intrinsics and the given pose. The semantic feature $\mathbf{f}_z \in \mathbb{R}^d$ is obtained through the semantic feature embedding process described previously. The class ID $y_z$ specifies the object category detected by YOLO, or is set to ``\texttt{null}'' for segments detected by FastSAM. The timestamp $t_z$ records when the observation is created.
This structured observation set serves as the input for the subsequent object-level map updating process. In the following section, we describe how the system incrementally organizes and maintains objects in the concrete map based on these observations.

\subsubsection{Scene Pointcloud Generation}

\begin{figure}[t]
    \centering
    \includegraphics[width=1\linewidth]{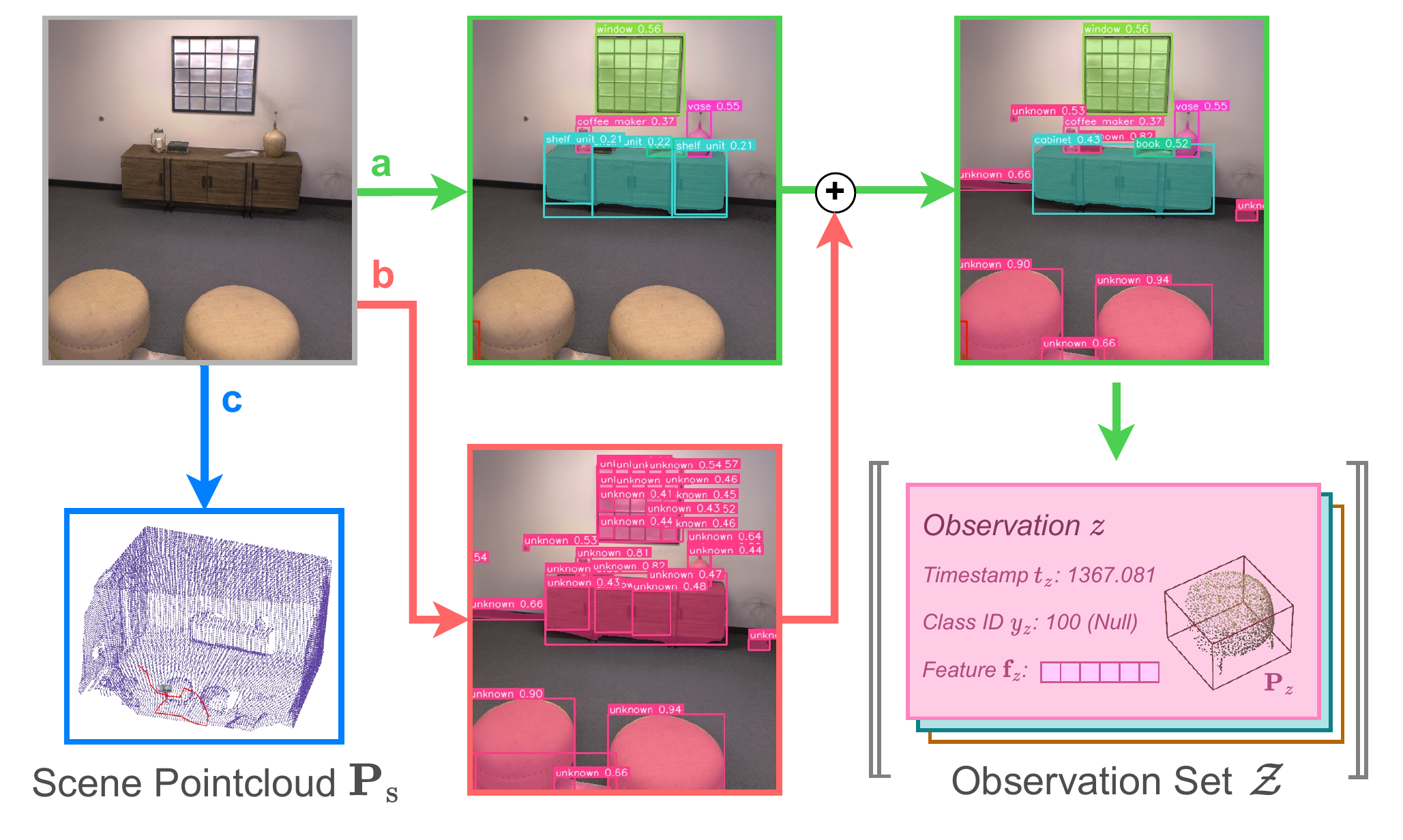}
    \caption{ Observation generation pipeline.
From each posed RGBD frame, the system runs three parallel threads:
a)~\textit{Closed-set detection} (e.g. YOLO) identifies in-list objects in object-level but misses unseen ones (e.g., \textit{stools});
b)~{\textit{Open-set segmentation}} (e.g. FastSAM) captures a broader object range but may over-segment (e.g., \textit{window}).
After refinement and merging, the results are embedded into the final observation set.
c)~{\textit{Scene cloud generation}} runs at low frequency and generates 3D point clouds for layout calculation.
    }
    \label{fig:obs}
    
    \vspace{-10pt}
    
\end{figure}

To support layout estimation for downstream navigation, we maintain a scene point cloud $\mathbf{P}_\text{s}$ that captures the overall geometry information of the environment during operation (Fig.~\ref{fig:obs}-c). It is built by selecting RGBD frames with significant pose changes (translation $>$ 1.0m or rotation $>$ 20°), projecting their point clouds, and applying downsampling. This process (blue branch in Fig.~\ref{fig:obs}) runs in a parallel low-frequency thread, providing a compact structural representation of the scene.

\subsection{Object Association and Update}
\subsubsection{Map Initialization}
The concrete map $\mathcal{M}_c$ represents a collection of objects, defined as $\mathcal{M}_c = \{ o_1, o_2, \ldots \}$. Each object $o$ is structured as $(\mathbf{P}_o, y_o, \mathbf{f}_o, L_o)$, where $\mathbf{P}_o$, $y_o$, and $\mathbf{f}_o$ inherit the same definitions as in the observation structure, and $L_o$ records the list of associated observations.
The initialization of the concrete map occurs when the first observation set $\mathcal{Z}_0$ is received. For each observation $z \in \mathcal{Z}_0$, a new object $o$ is created in the map as follows:
\begin{equation}
(\mathbf{P}_o, y_o, \mathbf{f}_o, L_o) = (\mathbf{P}_z, y_z, \mathbf{f}_z, \{ z \}).
\end{equation}

The initialized concrete map $\mathcal{M}_c$ sets up the basis for further incremental updates.
\subsubsection{Observation Matching}
When a new observation set $\mathcal{Z} = \{ z_1, z_2, \ldots, z_N \}$ arrives, we match the observations to the existing objects $\mathcal{M}_c = \{ o_1, o_2, \ldots, o_M \}$ in the concrete map. 
The matching relies on both the geometric representation $\mathbf{P}$ and the semantic feature $\mathbf{f}$ of each observation and object.
We define a similarity matrix $S \in \mathbb{R}^{N \times M}$, where each element $S(z_i, o_j)$ measures the similarity between observation $z_i$ and object $o_j$ as follows:
\begin{equation}
    S(z_i, o_j) = \cos(\mathbf{f}_{z_i}, \mathbf{f}_{o_j}) + \text{Overlap}(\mathbf{P}_{z_i}, \mathbf{P}_{o_j}),
\end{equation}
where $\cos(\cdot, \cdot)$ computes the cosine similarity between semantic features, and $\text{Overlap}(\cdot, \cdot)$ measures the 3D point cloud overlap ratio, following the calculation in \cite{gu2024conceptgraphs}.

For each observation $z_i$, we identify the best matching object based on the maximum similarity score:
\begin{equation}
o^* = \arg\max_{o_j} S(z_i, o_j).
\end{equation}
If the score $S(z_i, o^*)$ exceeds a threshold $\tau$, we associate $z_i$ with $o^*$ and update the object. We update $o^*$ by averaging $\mathbf{f}_{o^*}$ with $\mathbf{f}_{z_i}$ , weighted by the number of previous observations (i.e., the size of the observation list $L_{o^*}$), augmenting $\mathbf{P}_{o^*}$ with $\mathbf{P}_{z_i}$, and appending $z_i$ to $L_{o^*}$. Otherwise, a new object is created following the 
initialization procedure.

\begin{figure}[t!]
    \centering
    \includegraphics[width=1\linewidth]{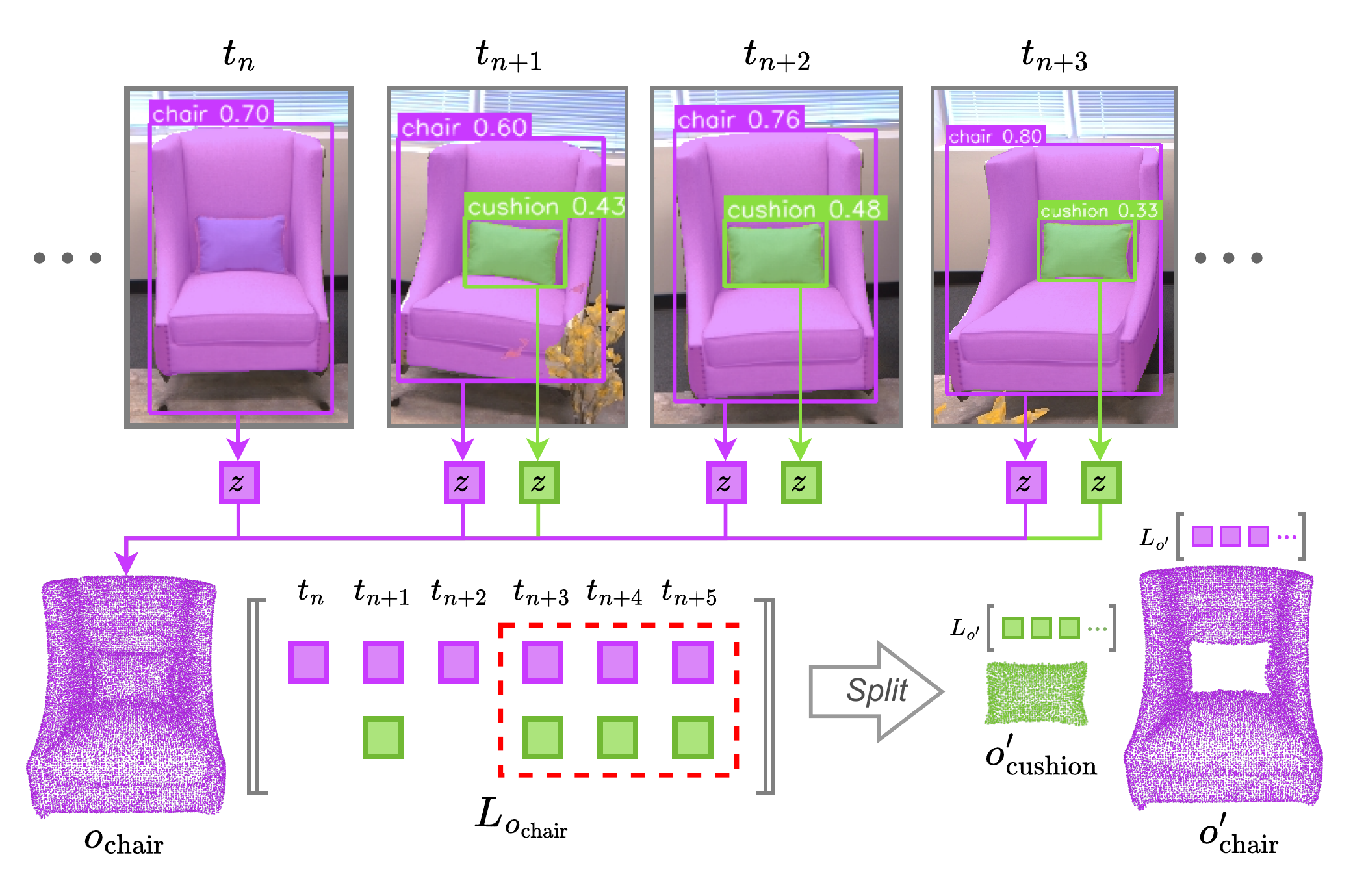}
    \caption{Split detection.
    At $t_n$, under-segmentation merges the cushion into the chair’s mask, producing an erroneous observation. Updating $o_\text{chair}$ with this observation corrupts its pointcloud, causing frequent misassignment of later cushion observations. The system detects such errors by checking $L_{o_\text{chair}}$ for co-occurring observations with different class IDs at the same timestamps. If such conflicts persist across frames (red box), a split is triggered to create new objects.
    }
    \label{fig:split}
    \vspace{-13pt}
\end{figure}

\subsubsection{Object Status Check}
\label{sec:status-check}
Our hybrid segmentation frontend produces holistic object observations. This avoids the need for expensive 3D inter-object merging, which typically combines fragmented segments into complete objects at high cost. Instead, after a map update with a new observation set $\mathcal{Z}$, we perform lightweight intra-object status checks—\textbf{stability check} and \textbf{split detection}—to maintain map fidelity. Both checks operate by analyzing the distribution of class IDs within an object's observation list $L_{o}$.

Stability check filters out insufficiently observed or noisy objects. An object triggers a stability check if it has not been updated for an extended period. It passes the check if it has accumulated a number of observations over the threshold and if its most frequent class ID accounts for at least two-thirds of the total observations in its observation list. Otherwise, it is considered unstable and removed from the map.

Although we have refined object-level segmentation results, under-segmentation can still occur, leading subsequent observations from adjacent objects to be incorrectly associated with a single object, as shown in Fig.~\ref{fig:split}. When observations with different class IDs appear at the same timestamp across consecutive frames, we trigger a split operation, partitioning the object's observation list by class IDs and creating new objects. This process preserves small objects and improves scene fidelity.

\section{Abstract Map and Navigation}
This section details how the abstract map $\mathcal{M}_a$ is constructed from the concrete map $\mathcal{M}_c$ (Fig.~\ref{fig:pipeline}-b), and how this dual-map design supports language-guided navigation under dynamic object changes (Fig.~\ref{fig:pipeline}-c).
We first describe the abstraction process, then explain how to retrieve navigation candidates from the abstract map given a language query. Next, we present a navigation strategy for static and dynamic environments. Finally, we show how the abstract map is updated online with a concrete map built during navigation.

\subsection{Map Abstraction}
\label{sec:map-abs}
The abstraction of the concrete map involves two steps: first, identifying which objects are selected as anchor objects; and second, associating the remaining volatile objects with anchor objects based on spatial relations. Anchor objects refer to static items that rarely move, such as beds or desks, while volatile objects (e.g., books or cups) frequently change positions. Only anchor objects are retained in the abstract map. 
We first classify objects using two lists of representative anchor and volatile categories.
For each object $o \in \mathcal{M}_c$, we compute its similarity to both lists using CLIP features.
If the maximum similarity to one list exceeds the other by a margin of $\Delta\tau = 0.05$, the object is assigned to the corresponding category.
For cases without a clear margin, we further compute the similarity between the object's feature $\mathbf{f}_o$ and the anchor template feature $\mathbf{f}_\text{t}$, encoded from descriptive phrases (e.g., ``furniture that is not often moved''). If the similarity exceeds a threshold $\tau_a$, the object is classified as an anchor object $a$; otherwise, it is a volatile $v$.
The threshold $\tau_a$ is set to 0.5 based on classification accuracy experiment.

During abstraction, all non-semantic attributes of volatile objects (e.g., $\mathbf{P}_v$, $L_v$) are discarded. 
We retain only the features of volatile objects that are spatially related to anchor objects. 
While our framework supports multiple relation types, we focus on the ``on'' relation in this work. To determine whether a volatile object is on an anchor, we first extract the anchor’s supporting plane by analyzing the Z-axis histogram of its point cloud. We then compute the 2D overlap ratio in the XY-plane to ensure the object lies within the anchor’s projected footprint, and verify that the vertical distance from the object’s bottom to the supporting plane is within a threshold $\delta$  (we set $\delta = 0.1$m in practice).
Once a spatial relation is established, the volatile object’s semantic feature $\mathbf{f}_{v}$ is stored in the associated feature list of the corresponding anchor object, denoted as $L_a = \{ \mathbf{f}_{v_1}, \mathbf{f}_{v_2}, \dots, \mathbf{f}_{v_K} \}$. Each anchor object $a \in \mathcal{M}_a$ is thus represented as $a = (\mathbf{P}_a, \mathbf{f}_a, y_a, L_a)$~\footnote{To support the navigation task, we simplify anchor object geometry $\mathbf{P}_a$ by preserving only its 2D projected point cloud.}. After abstraction, the abstract map $\mathcal{M}_a$ consists solely of anchor objects, and is represented as $\mathcal{M}_a = \{ a_1, a_2, \ldots \}$.

Scene layout, important for navigation, is also extracted in map abstraction. We project the scene point cloud $\mathbf{P}_\text{s}$ onto a bird’s-eye-view plane and discretize it into spatial bins. A binary occupancy map is then generated by computing a histogram of point counts across bins and using the 90th percentile as a threshold, highlighting scene structures like walls. 
The layout is visualized as gray point clouds in Fig.~\ref{fig:teaser}, and is updated only when a new navigation query is issued.

\subsection{Candidate Retrieval}
Given a natural language query $Q$, our goal is to retrieve the most relevant candidate from the abstract map $\mathcal{M}_a$ to serve as a navigation target. We first encode the query using CLIP to obtain a semantic embedding, denoted as $\mathbf{f}_q$.
We then compare the query feature $\mathbf{f}_q$ with both the anchor's own feature $\mathbf{f}_a$ and the features of its associated volatile objects $L_a = \{\mathbf{f}_{v_1}, \dots, \mathbf{f}_{v_K}\}$. We compute the anchor score as:
\begin{equation}
s(a) = \max\left( \cos(\mathbf{f}_q, \mathbf{f}_a),\ \max_i \cos(\mathbf{f}_q, \mathbf{f}_{v_i}) \right).
\end{equation}
We compute scores for all anchors in $\mathcal{M}_a$ and select the one with the highest score as the global navigation candidate: 
\begin{equation}
a^* = \arg\max_{a \in \mathcal{M}_a} s(a).
\end{equation}
The selected anchor $a^*$ suggests that the queried object is most likely situated nearby. Both the anchor $a^*$ and its similarity score $s(a^*)$ are used for further navigation.

\subsection{Navigation Strategy}

A global path toward ${a^*}$ is planned using a Voronoi-based planner over the abstract map~\cite{thrun1996integrating}. As the agent moves along this path, it incrementally builds a local concrete map $\mathcal{M}_c^{\text{local}}$ and evaluates objects in $\mathcal{M}_c^{\text{local}}$ for potential matches. For each object $o \in \mathcal{M}_c^{\text{local}}$, we compute the cosine similarity $s(o)$ between its feature $\mathbf{f}_o$ and the query feature $\mathbf{f}_q$. If $s(o)$ is within a margin $\epsilon$ of $s(a^*)$ and $o$ lies within the projected footprint of $a^*$, it is considered a confident match. A local path is then planned via RRT*~\cite{karaman2010incremental} to reach the target.

If no confident match is found near $a^*$, suggesting the queried object may have changed location, the system re-executes the candidate retrieval over the updated abstract map $\mathcal{M}_a' \setminus \{a^*\}$ and selects a new anchor $a^{*\prime}$. Here, $\mathcal{M}_a'$ is obtained by updating the original abstract map $\mathcal{M}_a$ with the local concrete map $\mathcal{M}_c^{\text{local}}$, as will be detailed in the next section. The agent then resumes navigation toward $a^{*\prime}$, using the original similarity score $s(a^*)$ to remain consistent with the initial query. This strategy enables the agent to leverage contextual cues encountered in the earlier navigation process, increasing the likelihood of success if the target object was partially observed along the way.

\subsection{Map Update Mechanism}

To update the abstract map $\mathcal{M}_a$, we extract stable objects from $\mathcal{M}_c^{\text{local}}$ and apply the same abstraction process to obtain new anchors $a^{\text{new}}$. Each $a^{\text{new}}$ is compared against all existing anchors $a \in \mathcal{M}_a$ by computing the geometric overlap score $s_{\text{overlap}} = \text{Overlap}(\mathbf{P}_{a^{\text{new}}}, \mathbf{P}_a)$.
If $s_{\text{overlap}} > \tau_1$, we treat $a^{\text{new}}$ as an additional observation of $a$ and update its representation. The point cloud $\mathbf{P}_a$ is augmented with $\mathbf{P}_{a^{\text{new}}}$ and downsampled, and the semantic feature is updated using a size-weighted average:
\begin{equation}
\mathbf{f}_a \leftarrow \frac{|\mathbf{P}_a| \cdot \mathbf{f}_a + |\mathbf{P}_{a^{\text{new}}}| \cdot \mathbf{f}_{a^{\text{new}}}}{|\mathbf{P}_a| + |\mathbf{P}_{a^{\text{new}}}|}.
\end{equation}
The symbol $|\mathbf{P}|$ refers to the size of the pointcloud $\mathbf{P}$.
If the overlap score further exceeds a stricter threshold $\tau_2$, indicating that $a^{\text{new}}$ provides a more complete observation of the existing anchor, we additionally replace the volatile feature list $L_a$ with $L_{a^{\text{new}}}$. 
If no anchor satisfies $s_{\text{overlap}} > \tau_1$, we insert $a^{\text{new}}$ into the abstract map as a new anchor.
The updated abstract map $\mathcal{M}_a'$ maintains the most up-to-date scene representation throughout navigation.

\section{Experiments}
\label{exp}

This section first describes the experimental setup and further evaluates DualMap across four core aspects: 1) the effectiveness and efficiency of online concrete map construction; 2) the navigation performance in both static and dynamic scenes; 3) the contribution of key system components (via ablations); and 4) real-world applicability.

\subsection{Benchmarks and Implementation Details}
\subsubsection{Datasets}
\label{sec:datasets}
We evaluate DualMap on semantic segmentation, efficiency, and object navigation.
For semantic segmentation and efficiency, we use the ScanNet~\cite{dai2017scannet} and Replica~\cite{straub2019replica} datasets, following scene selections from prior work~\cite{gu2024conceptgraphs, werby23hovsg}, with all experiments conducted at the original image resolutions.
For static object navigation, we evaluate on the HM3D dataset~\cite{yadav2022habitat}, selecting three representative scenes (\texttt{00829}, \texttt{00848}, \texttt{00880}) and augmenting object diversity by integrating YCB objects~\cite{calli2017yale}.
For dynamic navigation, as current benchmarks lack object-level dynamics, we construct new environments for dynamic object navigation using a custom tool built on Habitat Simulator~\cite{puig2023habitat3}, which enables dynamic placement of YCB objects.  
We define two types of object changes: 
\textit{In-anchor relocation}, where an object moves within the same anchor (e.g., moving a cup on a table);  
\textit{Cross-anchor relocation}, where an object is moved to a different anchor (e.g., from a table to a shelf).  
For each static HM3D scene, we create dynamic variants by manually repositioning added objects to reflect both types of changes.  
Agents are deployed in the Habitat Simulator for both static and dynamic navigation experiments.
For real-world experiments, we design a rigid perception module with a LiDAR and an RGBD camera, where LiDAR provides pose estimation via FastLIO2~\cite{xu2022fast}. This module is deployed on both wheeled and quadruped platforms, as shown in Fig.~\ref{fig:real-world}-a.

\begin{figure*}[t!]
    \centering
    \includegraphics[width=1.0\textwidth]{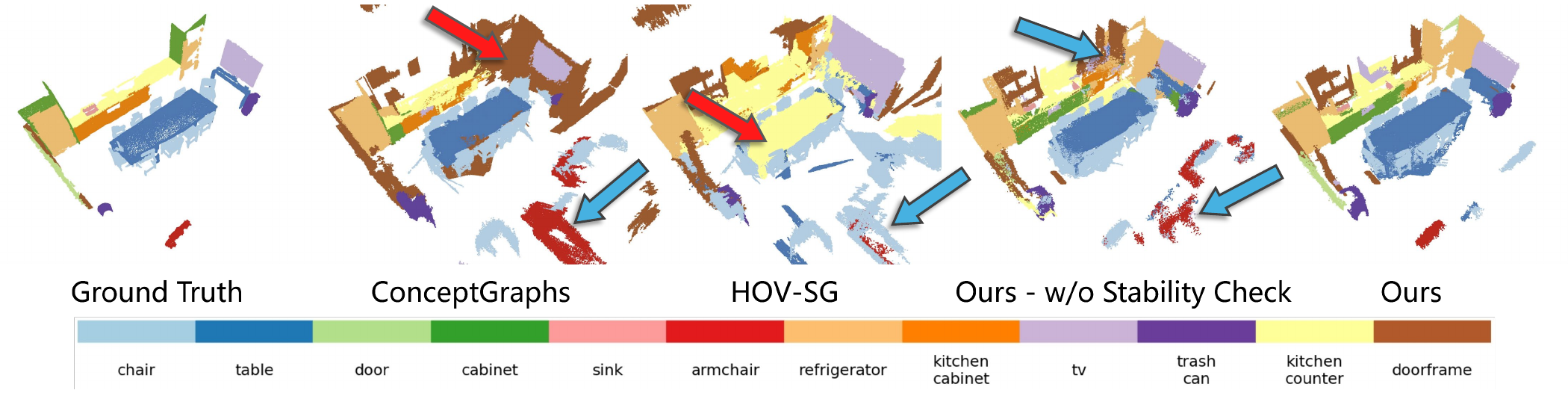}
    \caption{Qualitative comparison of semantic segmentation results on ScanNet \texttt{scene0011\_00}. \textcolor[HTML]{FF1919}{Red} arrows highlight semantically inaccurate predictions, while  \textcolor[HTML]{46B1E1}{blue} arrows indicate meaningless segmentations, suggesting noisy predictions.}
    \label{fig:qualitative}
    \vspace{-5pt}
\end{figure*}

\begin{table*}[t!]
\centering
\caption{Open-vocabulary 3D Semantic Segmentation and Efficiency}
\label{tab:concretemap}
\resizebox{0.98\textwidth}{!}{%
\begin{tabular}{cc ccccccc}
\toprule
\textbf{Dataset} & \textbf{Method} & \textbf{mIoU $\uparrow$} & \textbf{FmIoU $\uparrow$} & \textbf{mAcc $\uparrow$} & \textbf{ODR $\approx 1$} & \textbf{Avg. Mem (MB) $\downarrow$} & \textbf{Peak Mem (MB) $\downarrow$} & \textbf{TPF (s) $\downarrow$} \\
\midrule
\multirow{3}{*}{Replica}
& ConceptGraphs & 0.1501 & 0.3858 & 0.3559 & 2.02 & 7148.9 & 23551.9 & 4.188 \\
& HOV-SG        & 0.2050 & 0.4846 & 0.3835 & 3.81 & 73368.0 & 158126.6 & 42.005 \\
& \textbf{Ours} & \textbf{0.2538} & \textbf{0.5207} & \textbf{0.4024} & \textbf{0.97} & \textbf{3095.2} & \textbf{4564.0} & \textbf{0.276} \\
\midrule
\multirow{3}{*}{ScanNet}
& ConceptGraphs & 0.0882 & 0.3077 & 0.3538 & 6.97 & 9780.3 & 26155.2 & 6.301 \\
& HOV-SG        & 0.1333 & \textbf{0.3381} & 0.3714 & 20.34 & 9223.0 & 25735.0 & 8.039 \\
& \textbf{Ours} & \textbf{0.1604} & 0.3288 & \textbf{0.3794} & \textbf{2.56} & \textbf{2120.9} & \textbf{2820.2} & \textbf{0.163} \\
\bottomrule
\end{tabular}%
}
\vspace{-8pt}
\end{table*}

\subsubsection{Evaluation Metrics}
For semantic segmentation, we adopt standard metrics from~\cite{werby23hovsg}, including mean IoU (mIoU), frequency-weighted IoU (F-mIoU), and mean accuracy (mAcc). 
In both qualitative and quantitative evaluations, we exclude background classes like \texttt{["wall", "floor", "ceiling"]}, which occupy large portions of the scene but provide little semantic information.
We also report the Object Density Ratio (ODR), a novel metric defined as the ratio of predicted object counts to ground-truth object counts. This metric captures how well the system reproduces realistic object quantities; values closer to 1 indicate better alignment with real-world object density. ODR complements existing segmentation metrics by revealing over- or under-segmentation issues that may be hidden under high pixel-wise accuracy.
For efficiency, we measure average memory usage, peak memory consumption, and time per frame (TPF) during operation.
For navigation, we report Success Rate (SR), defined as the percentage of queries in which the agent stops within 1 meter of the queried object. In dynamic scenes, success further requires finding the target within three attempts.
Unless otherwise specified, all reported metric values in tables are averaged across scenes for each dataset.

\subsubsection{Compared Methods}
We compare against two competitive open-vocabulary semantic mapping systems: ConceptGraphs~\cite{gu2024conceptgraphs}, which supports online object-level mapping, and HOV-SG~\cite{werby23hovsg}, known for strong semantic performance and built-in navigation ability.
We extend both baselines in Habitat Simulator by enabling agents to navigate using their query results and the same planning algorithm as ours.

\subsubsection{Implementation Details}
All experiments are conducted on a machine with an NVIDIA RTX 4090 GPU and an Intel i7-12700KF CPU.
We adopt the YOLOv8l-world model~\cite{cheng2024yolo} for closed-set detection, using a class list generated by GPT-4~\cite{achiam2023gpt}, and supplement it with FastSAM-s~\cite{zhao2023fast} to enable open-vocabulary segmentation. For feature embedding, we use MobileCLIP-S2~\cite{mobileclip2024}, shared across all methods for fair comparison.

\subsection{Concrete Mapping Evaluation}

We present quantitative and qualitative results in Table~\ref{tab:concretemap} and Fig.~\ref{fig:qualitative}.
DualMap achieves the best semantic segmentation results across datasets, improving mIoU on Replica by 2.8\% over HOV-SG and 10.3\% over ConceptGraphs. This gain stems from object-level splitting that preserves the diversity of objects and text-enhanced feature embedding. Fig.~\ref{fig:qualitative} further illustrates these improvements, with red arrows showing misclassified objects that are better handled by DualMap.
For ODR, DualMap achieves the closest alignment with ground truth (0.97 on Replica), thanks to holistic segmentation that reduces over-segmentation and our stability check that removes noisy objects. While HOV-SG and ConceptGraphs apply inter-object merging to control segment counts, they still suffer from redundant fragments. This difference is clearly illustrated in Fig.~\ref{fig:qualitative}, where noisy RGBD inputs from real sensors result in meaningless predictions (blue arrows) in the baselines. DualMap filters out such errors via stability check, demonstrating robustness in real-world scenarios.

In efficiency, DualMap also outperforms other methods: it reduces peak memory usage by over \textbf{96\%} and TPF by \textbf{99.3\%} compared to HOV-SG on Replica. This is enabled by a hybrid open-vocabulary detection strategy and lightweight 2D refinement, which avoids costly 3D post-processing.

\begin{figure*}[t]
    \centering
    \includegraphics[width=1\linewidth]{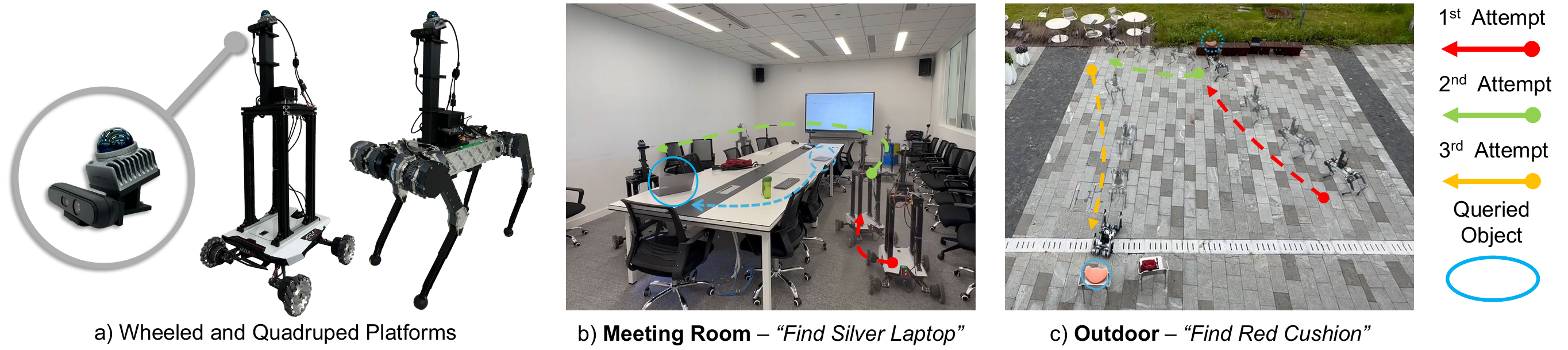}
    \caption{
    Real-world navigation in dynamic environments.
    a) Robotic platforms equipped with a perception module that integrates a LiDAR and an RGB-D camera, both mounted on a rigid 3D-printed mount.
    b–c) Two examples of language-guided navigation in dynamic real-world scenes, where the agent tries to locate relocated objects across multiple attempts.}
   
    \label{fig:real-world}
    \vspace{-10pt}
\end{figure*}

\subsection{Navigation in Simulated Environments}

\begin{table}[t]
\centering
\caption{Object Navigation Success Rate (SR) on HM3D.}
\label{tab:navigation}
\renewcommand{\arraystretch}{1.2}
\resizebox{\columnwidth}{!}{%
\begin{tabular}{cccccccc}
\toprule
\textbf{Scene Type} & \textbf{Method}             & \textbf{00829} & \textbf{00848} & \textbf{00880} & \textbf{Trials} & \textbf{Avg. SR} \\ \midrule
\multirow{3}{*}{Static}  
           & ConceptGraphs      & \underline{69.2\%} & \underline{53.8\%} & \underline{61.5\%} & \multirow{3}{*}{78}  & \underline{61.5\%} \\
           & HOV-SG             & 53.8\%              & 46.2\%              & 57.7\%              &  & 52.6\% \\
           & Ours      & \textbf{73.1\%}     & \textbf{69.2\%}     & \textbf{69.2\%}     &  & \textbf{70.5\%} \\ \midrule
\multirow{2}{*}{Dynamic}  
           & In-anchor (Ours)   & 66.7\%              & 66.7\%              & 61.1\%              & 54 & 64.8\% \\
           & Cross-anchor (Ours)& 55.6\%              & 61.1\%              & 64.7\%              & 53 & 60.3\% \\ \bottomrule
\end{tabular}%
}
\vspace{-10pt}
\end{table}

For static navigation, we evaluate all methods across three HM3D scenes. As shown in Table~\ref{tab:navigation}-Static, DualMap achieves the highest success rate. ConceptGraphs performs competitively due to its object-centric mapping. In contrast, HOV-SG's segment-merging strategy still results in fragmented maps, leading to more query and navigation failures.

For dynamic scenes (Table~\ref{tab:navigation}-Dynamic), DualMap naturally handles \textit{in-anchor} changes, as the queried object remains within its original anchor and can be accurately localized through the local concrete map.  For more challenging \textit{cross-anchor} cases, it updates the abstract map online and reselects reasonable candidates accordingly. This strategy maintains high success rates even under large object movements. Among failure cases in \textit{cross-anchor} setting, 28.3\% are due to false matches, 9.4\% to exceeding navigation attempt limits, and 1.9\% to planning errors.

\subsection{Ablation Study}
We perform ablations to validate the impact of key components in DualMap, focusing on semantic mapping, efficiency, and navigation under dynamic scenes.
\subsubsection{Mapping Performance}

\begin{table}[t!]
\centering
\caption{Semantic Mapping Ablation on Replica}
\label{tab:abalation-seg}
\begin{tabular}{@{}ccccc@{}}
\toprule
\multicolumn{1}{c}{\textbf{Options}}                  & \textbf{FmIoU}           & \textbf{mAcc}            & \textbf{mIoU}    \\ \midrule
\multicolumn{1}{c}{w/o FastSAM}              & 0.4753          & 0.3685          & 0.2344          \\
\multicolumn{1}{c}{w/o YOLO Refinement}     & 0.5043          & 0.3886          & 0.2399                 \\ 
\multicolumn{1}{c}{w/o Weighted Feats}       & 0.4209          & 0.3348          & 0.1814  \\
\multicolumn{1}{c}{w/o Object Split Detection}         & 0.5065          & 0.3998          & 0.2496            \\ \midrule
\multicolumn{1}{c}{Full System}                      & {\textbf{0.5207}}    & {\textbf{0.4024}}    & \textbf{0.2538}         \\ \bottomrule
\end{tabular}

\vspace{-8pt}

\end{table}

In Table~\ref{tab:abalation-seg} and Fig.~\ref{fig:qualitative}, we ablate key components of the concrete mapping pipeline on Replica and ScanNet.
We first remove FastSAM and rely only on YOLO for detection. Since YOLO detects only predefined categories, many unseen objects are missed, leading to a drop in performance.
Next, we disable the refinement step that merges overlapping YOLO detections based on color histogram similarity. Without merging overlaps, objects are over-segmented. The performance drops slightly, as YOLO produces relatively holistic detections.
Then, we remove the weighted feature merging and use only image features. This causes a notable drop, indicating the importance of incorporating global semantic cues from text.
Finally, we disable the object-level splitting mechanism. This causes under-segmentation, where nearby objects are grouped into one, leading to inaccurate matching and reduced mapping precision.
Qualitative results in Fig.~\ref{fig:qualitative} show that without object stability check, sensor noise introduces noisy segments, degrading overall semantic quality.

\subsubsection{Efficiency}

\begin{table}[t]
\caption{Mapping Efficiency Ablation on ScanNet. Gray row indicates configuration used in our proposed system.}
\label{tab:ablation-effi}
\centering
\resizebox{\columnwidth}{!}{
\begin{tabular}{ccccc}
\toprule
\textbf{Obj. Merging} & \textbf{Stability Check} & \textbf{Closed-set Det.} & \textbf{ODR} & \textbf{TPF(s)} \\ \midrule
\ding{51} & \ding{55} & \ding{55} & 34.2 & 27.04  \\
\ding{51} & \ding{55} & \ding{51} & 26.5 & 2.841  \\
\ding{51} & \ding{51} & \ding{51} & \textbf{2.41} & 0.3056 \\
\rowcolor{gray!20} \ding{55} & \ding{51} & \ding{51} & 2.56 & \textbf{0.163} \\ \bottomrule
\end{tabular}
}
\end{table}

In Table~\ref{tab:ablation-effi}, we analyze three key modules that are related to our mapping pipeline to evaluate their impact on efficiency: Inter-object merging in 3D, intra-object level stability check, and closed-set detection assistance in open-vocabulary segmentation.
HOV-SG follows the configuration in Row 1, resulting in the highest time cost and ODR. This is because SAM-style segmentation produces a large number of segments, making 3D-level merging computationally expensive. Configurations in Row 2 improve speed by using YOLO to assist segmentation, but still suffer from noisy objects caused by sensor noise. In Row 3, enabling our proposed stability check significantly improves ODR by removing noisy objects, which also reduces the object merging time cost. Our final configuration (Row 4) disables inter-object merging, slightly sacrificing map accuracy but achieving notable efficiency gains—an important trade-off for real-world robotic applications.

\subsubsection{Dynamic Scene Navigation}
\begin{table}[t]
\caption{Success Rates under Different Candidate Selection Strategies for Relocated Objects on HM3D}
\label{tab:ablation-nav}
\centering
\begin{tabular}{@{}lccc@{}}
\toprule
\textbf{Strategy} & Random Pick & Based on $\mathcal{M}_a$ & \textbf{Based on $\mathcal{M}_a'$} \\
\midrule
\textbf{SR} & 13.2\% & 47.2\% & \textbf{60.3\%} \\
\bottomrule
\end{tabular}
\vspace{-15pt}
\end{table}

To evaluate navigation under dynamic changes, we focus on the more challenging \textit{cross-anchor} setting and test different candidate selection strategies when the queried object has changed position. As shown in Table~\ref{tab:ablation-nav}, random selection often fails due to excessive retries. Using the original abstract map $\mathcal{M}_a$ provides anchor-based guidance and improves success. Our final strategy updates $\mathcal{M}_a$ during navigation to form $\mathcal{M}_a'$, which makes full use of past navigation information. This helps the agent find the relocated object more effectively.

\subsection{Navigation in Real-World Environments} 

\begin{table}[t]
\centering
\caption{Real-World Object Navigation Results}
\label{tab:realworld}
\begin{tabular}{llcccc}
\toprule
\multirow{2}{*}{\textbf{Platform}} & \multirow{2}{*}{\textbf{Scene Type}} & \multicolumn{2}{c}{\textbf{Static}} & \multicolumn{2}{c}{\textbf{Dynamic}} \\
 & & \textbf{Trials} & \textbf{SR} & \textbf{Trials} & \textbf{SR} \\
\midrule
\multirow{2}{*}{Wheeled} 
  & Meeting Room     & 14 & 85.7\% & 27 & 70.3\% \\
  & Apartment        & 46 & 69.6\% & 33 & 51.5\% \\
\midrule
\multirow{2}{*}{Quadruped} 
  & Indoor Hallway   & 19 & 78.9\% & 27 & 55.6\% \\
  & Outdoor          & 12 & 75.0\% & 18 & 50.0\% \\
\bottomrule
\end{tabular}
\vspace{-16pt}
\end{table}

We evaluate our system on two robotic platforms across four diverse real-world scenes, as shown in Table~\ref{tab:realworld} and Fig.~\ref{fig:real-world}.
To construct test environments, we place diverse objects (e.g., bottles, backpacks, keyboards) within each scene. During mapping, the robot is manually driven to explore and build an abstract map. We then perform static object navigation using the same pipeline as in simulation. To test dynamic scenarios, we manually rearrange the objects and initiate navigation queries. The results show that DualMap achieves performance levels similar to those in the simulations, confirming its robustness in real-world settings across scenes and platforms.

\section{Conclusion}
\label{conclusion}
We present DualMap, an online open-vocabulary semantic mapping system for language-guided navigation in dynamic environments. The system enables efficient and accurate mapping, and supports robust navigation through online updates and candidate re-selection.
While effective, our system has several limitations. First, DualMap does not perform camera pose estimation and instead relies on external localization systems. 
Second, the system does not yet incorporate human-object interactions, limiting its ability to reason about human activities in the scene. 
Third, even in simple outdoor scenes, system performance degrades compared to indoor settings due to higher sensor noise.
We consider these limitations promising directions for future work.

\bibliographystyle{IEEEtran}
\bibliography{ref}

\newpage
\appendix
\appendices

\section{Prompt Example}

\label{supp-prompt}
To incorporate prior knowledge of the robot's operating environment, we utilize GPT-4o to generate a potential list of detection classes. The prompt and the returned class list are as follows:

\begin{tcolorbox}[colback=gray!10,title={User Prompt}]
I am deploying a robot in an indoor house setting and the robot needs a list of object categories to detect. 
Based on class name formats from the ScanNet and Replica datasets, provide 100 common indoor categories from living rooms, bedrooms, restrooms, kitchens, etc. 
The last category should be `null'.
\end{tcolorbox}

\begin{tcolorbox}[colback=blue!5,title={GPT-4o Output}]
chair, sofa, table, bed, desk, cabinet, shelf, book, lamp, tv, television stand, refrigerator, microwave, oven, sink, washing machine, dryer, stove, dishwasher, toaster, kettle, coffee maker, towel, toilet, shower, bathtub, mirror, picture frame, door, window, floor lamp, ceiling lamp, fan, air conditioner, heater, carpet, rug, trash can, recycling bin, iron, ironing board, broom, vacuum cleaner, trash bag, laundry basket, tissues, soap dispenser, toilet paper, toothbrush, toothpaste, hair dryer, shampoo, conditioner, shampoo dispenser, toilet brush, shampoo bottle, soap bar, candle, plant, pot, vase, clock, speaker, remote control, fan heater, curtains, blinds, curtain rod, frame, desk chair, bed frame, dresser, nightstand, wardrobe, closet, pillows, blanket, mattress, pillowcase, slippers, shoes, hangers, coat rack, wall, floor, ceiling, staircase, hallway, entryway, shelf unit, drawer, light switch, outlet, plug, router, computer, monitor, keyboard, mouse, printer, null.
\end{tcolorbox}

This output class list, named \texttt{gpt\_indoor\_general} in the released code, was used for our indoor experiments.

\section{Detailed Experiment Setup}
\label{supp-dataset}

\subsection{Datasets}
For ScanNet, we use the following scenes: \texttt{scene0011\_00}, \texttt{scene0050\_00}, \texttt{scene0231\_00}, \texttt{scene0378\_00}, and \texttt{scene0518\_00}. 
We adopt the ScanNet200 benchmark, which shares the same geometry as ScanNet20 but provides a broader set of class categories and more detailed semantic annotations for evaluating 3D scene understanding methods. 
For Replica, the evaluated scenes are: \texttt{office0-office4} and \texttt{room0-room2}. 
To evaluate query performance across a wider range of semantic categories, we use HM3DSem, specifically the scenes \texttt{00829}, \texttt{00848}, and \texttt{00880}.
To enrich the object categories, we place six objects from the YCB dataset at random positions within each HM3DSem scene. The object models used are: \texttt{003\_cracker\_box}, \texttt{005\_tomato\_soup\_can}, \texttt{011\_banana}, \texttt{019\_pitcher\_base}, \texttt{024\_bowl}, \texttt{025\_mug}, \texttt{029\_plate}, and \texttt{037\_scissors}.

\subsection{Metric Definitions}
In semantic segmentation evaluation, three metrics mIoU, FmIoU, mAcc are defined as follows:

\begin{equation}
\text{mIoU} = \frac{1}{C} \sum_{i=1}^{C} \frac{\text{TP}_i}{\text{TP}_i + \text{FN}_i + \text{FP}_i}
\end{equation}

\begin{equation}
\text{FmIoU} = \sum_{i=1}^{C} \frac{\text{TP}_i + \text{FN}_i}{\sum_{j=1}^{C} (\text{TP}_j + \text{FN}_j)} \times \frac{\text{TP}_i}{\text{TP}_i + \text{FN}_i + \text{FP}_i}
\end{equation}

\begin{equation}
\text{mAcc} = \frac{1}{C} \sum_{i=1}^{C} \frac{\text{TP}_i}{\text{TP}_i + \text{FP}_i}
\end{equation}

where $C$ is the number of classes, and $\text{TP}_i$, $\text{FP}_i$, and $\text{FN}_i$ denote the true positives, false positives, and false negatives for class $i$, respectively. In FmIoU, the first term represents the class frequency, while the second is the IoU of class $i$.

\subsection{List of Navigation Tasks}
In the HM3DSem object navigation experiments, we selected representative objects from the ground truth labels, covering a variety of common categories found in indoor environments. The complete list of objects used for the navigation tasks is provided in Table \ref{tab:query-text}. The added YCB objects are marked with \underline{underline}, and Fig.~\ref{fig:dynamic-examples}-a illustrates the original placement of these YCB objects across the test scenes.

\begin{table}[t]
\caption{Queried Objects in HM3D Test}
\centering
\label{tab:query-text}
\begin{tabular}{cp{7cm}}
\toprule
\textbf{Scene} & \textbf{Objects for Navigation Evaluation } \\ \midrule
\texttt{00829} & chair, picture, towel, table, ottoman, tap, sofa, bin, tv, cabinet, magazine, washbasin counter, bed, telephone, clothes, bathtub, bag, tv stand, decoration, toilet, \underline{cracker box}, \underline{soup can}, \underline{pitcher}, \underline{bowl}, \underline{plate}, \underline{scissors}                                           \\ \midrule
\texttt{00848} & pillow, tap, stool, toilet paper, towel, magazine, vase, bed, armchair, bowl of fruit, bathroom counter, christmas tree, bench, kettle, coffee maker, microwave, cooker, refrigerator, kitchen island, bathtub, \underline{cracker box}, \underline{soup can}, \underline{pitcher}, \underline{mug}, \underline{plate}, \underline{scissors}, \underline{banana} \\ \midrule
\texttt{00880} & shelf, painting, table, cabinet, curtain, container, mirror, hat, tv, washing machine, laptop, clothes, couch, microwave, sink, trashcan, dishwasher, ironing board, printer, desk, \underline{cracker box}, \underline{soup can}, \underline{pitcher}, \underline{bowl}, \underline{plate}, \underline{scissors} \\ \bottomrule
\end{tabular}
\end{table}

\subsection{Dynamic Setting}
\begin{figure*}[t]
    \centering
    \includegraphics[width=1\linewidth]{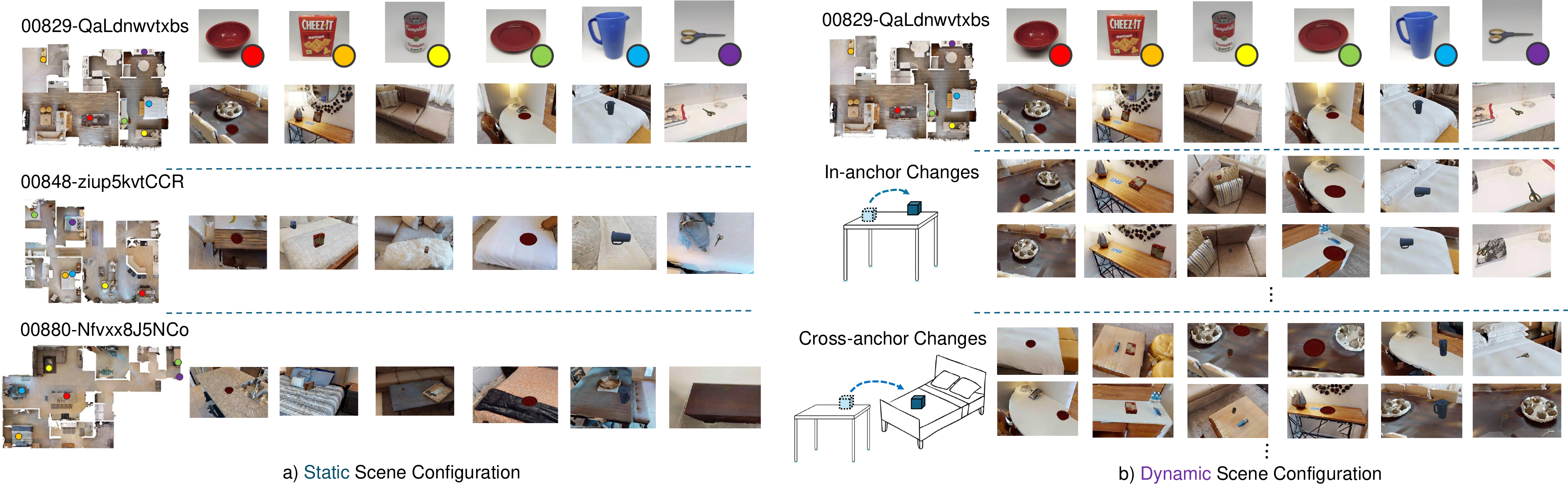}
    \caption{Examples of YCB object configurations: a) YCB objects are manually placed across three HM3D scenes. b) The YCB objects are dynamically relocated via two dynamic change types for the navigation test.}
    \label{fig:dynamic-examples}
    \vspace{-10pt}
\end{figure*}

The added YCB dataset objects are also used in dynamic change query experiments. Following the two types of dynamic changes defined in Sec.\ref{sec:datasets}, each object is moved three times for both\textit{ in-anchor} and \textit{cross-anchor} changes, showing in Fig.~\ref{fig:dynamic-examples}.
To facilitate reproducibility, detailed experiment results are presented in Tables~\ref{tab:cross_anchor_experiment_data_phase2} and~\ref{tab:cross_anchor_experiment_data}.

\begin{table}[t!]
\centering
\caption{Detailed In-Anchor Experiment Results}
\label{tab:cross_anchor_experiment_data_phase2}
\resizebox{\columnwidth}{!}{%
\begin{tabular}{lclclc}
\toprule
\multicolumn{2}{c}{\textbf{00829}} & \multicolumn{2}{c}{\textbf{00848}} & \multicolumn{2}{c}{\textbf{00880}} \\
\cmidrule(lr){1-2} \cmidrule(lr){3-4} \cmidrule(lr){5-6}
Type & Success & Type & Success & Type & Success \\
\midrule
\multicolumn{6}{l}{\textit{\textbf{0116.json}}} \\
\midrule
soup\_can    & 1 & scissors     & 1 & soup\_can    & 1 \\
bowl         & 1 & plate        & 1 & bowl         & 1 \\
plate        & 1 & pitcher      & 0 & pitcher      & 1 \\
scissors     & 0 & mug          & 1 & plate        & 0 \\
cracker\_box & 1 & banana       & 1 & cracker\_box & 0 \\
pitcher      & 1 & cracker\_box & 1 & scissors     & 0 \\
\midrule
\multicolumn{6}{l}{\textit{\textbf{0117.json}}} \\
\midrule
soup\_can    & 1 & scissors     & 1 & soup\_can    & 1 \\
bowl         & 1 & plate        & 1 & bowl         & 1 \\
plate        & 1 & pitcher      & 0 & pitcher      & 1 \\
scissors     & 0 & mug          & 0 & plate        & 0 \\
cracker\_box & 1 & banana       & 1 & cracker\_box & 0 \\
pitcher      & 0 & cracker\_box & 1 & scissors     & 1 \\
\midrule
\multicolumn{6}{l}{\textit{\textbf{0118.json}}} \\
\midrule
soup\_can    & 0 & scissors     & 0 & soup\_can    & 1 \\
bowl         & 1 & plate        & 1 & bowl         & 1 \\
plate        & 1 & pitcher      & 0 & pitcher      & 1 \\
scissors     & 0 & mug          & 0 & plate        & 0 \\
cracker\_box & 1 & banana       & 1 & cracker\_box & 0 \\
pitcher      & 0 & cracker\_box & 1 & scissors     & 1 \\
\midrule
\multicolumn{6}{c}{\textbf{Summary}} \\
\midrule
\textbf{Total} & \textbf{12/18} & \textbf{Total} & \textbf{12/18} & \textbf{Total} & \textbf{11/18} \\
\bottomrule
\end{tabular}%
}
\end{table}

\begin{table}[t!]
\centering
\caption{Detailed Cross-Anchor Experiment Results}
\label{tab:cross_anchor_experiment_data}
\resizebox{\columnwidth}{!}{%
\begin{tabular}{lclclc}
\toprule
\multicolumn{2}{c}{\textbf{00829}} & \multicolumn{2}{c}{\textbf{00848}} & \multicolumn{2}{c}{\textbf{00880}} \\
\cmidrule(lr){1-2} \cmidrule(lr){3-4} \cmidrule(lr){5-6}
Type & Success & Type & Success & Type & Success \\
\midrule
\multicolumn{6}{l}{\textit{\textbf{0129-1.json}}} \\
\midrule
bowl        & 1 & cracker\_box & 1 & soup\_can    & 1 \\
cracker\_box& 0 & scissors     & 1 & pitcher      & 0 \\
soup\_can   & 1 & mug          & 0 & bowl         & 1 \\
plate       & 1 & pitcher      & 1 & plate        & 0 \\
pitcher     & 1 & plate        & 1 & scissors     & 1 \\
scissors    & 0 & banana       & 0 & cracker\_box  & 1 \\
\midrule
\multicolumn{6}{l}{\textit{\textbf{0128-2.json}}} \\
\midrule
bowl        & 1 & cracker\_box & 1 & soup\_can    & 1 \\
cracker\_box& 0 & scissors     & 1  & pitcher      & 0 \\
soup\_can   & 0 & mug          & 0 & bowl         & 1 \\
plate       & 0 & pitcher      & 0 & plate        & 1 \\
pitcher     & 1 & plate        & 1 & scissors     & 0 \\
scissors    & 1 & banana       & 1 &              &   \\
\midrule
\multicolumn{6}{l}{\textit{\textbf{0128-1.json}}} \\
\midrule
bowl        & 1 & cracker\_box & 0 & soup\_can    & 0 \\
cracker\_box& 0 & scissors     & 1 & pitcher      & 1 \\
soup\_can   & 1 & mug          & 1 & bowl         & 1 \\
plate       & 1 & pitcher      & 0 & plate        & 1 \\
pitcher     & 0 & scissors     & 1 & scissors     & 0 \\
scissors    & 0 & banana       & 1 & cracker\_box & 0 \\
\midrule
\multicolumn{6}{c}{\textbf{Summary}} \\ 
\midrule
\textbf{Total} & \textbf{10/18} & \textbf{Total} & \textbf{12/18} & \textbf{Total} & \textbf{10/17} \\
\bottomrule
\end{tabular}%
}
\end{table}

\section{Experiments For Threshold}
In this section, we present the experiments conducted to determine the threshold, as discussed in Sec.~\ref{sec:feat-weight} and Sec.~\ref{sec:map-abs}.

\subsection{Feature Weights}

\begin{table}[t!]
\centering
\caption{Ablation Study on Feature Weighting}
\vspace{5pt}
\label{tab:feat-weight-ablation}
\resizebox{\columnwidth}{!}{
\begin{tabular}{ccccc|ccccc}
\toprule
\multicolumn{5}{c|}{\textbf{Replica}}                           & \multicolumn{5}{c}{\textbf{Scannet}}                            \\
 \textbf{$\mathbf{f}_{\text{image}}$} & \textbf{$\mathbf{f}_{\text{text}}$} & \textbf{FmIoU} & \textbf{mAcc}  & \textbf{mIoU}  & $\mathbf{f}_{\text{image}}$ & $\mathbf{f}_{\text{text}}$ & \textbf{FmIoU} & \textbf{mAcc}  & \textbf{mIoU}  \\ \midrule
0.0   & 1.0  & 0.531          & 0.388          & 0.237          & 0.0   & 1.0  & 0.284          & 0.318          & 0.140          \\
0.1   & 0.9  & 0.532          & 0.384          & 0.239          & 0.1   & 0.9  & 0.309          & 0.328          & 0.150          \\
0.3   & 0.7  & 0.545          & 0.420          & 0.249          & 0.3   & 0.7  & 0.305          & 0.350          & 0.152          \\
0.5   & 0.5  & \textbf{0.557} & \textbf{0.439} & \textbf{0.262} & 0.5   & 0.5  & 0.321          & 0.365          & \underline{0.163}    \\
\rowcolor{gray!20}0.7   & 0.3  & \underline{0.551}    & \underline{0.425}    & \underline{0.251}    & 0.7   & 0.3  & \textbf{0.334} & \textbf{0.371} & \textbf{0.167} \\
0.9   & 0.1  & 0.498          & 0.380          & 0.201          & 0.9   & 0.1  & \underline{0.323}    & \underline{0.368}    & 0.158          \\
1.0   & 0.0  & 0.430          & 0.358          & 0.174          & 1.0   & 0.0  & 0.301          & 0.356          & 0.154          \\ \bottomrule
\end{tabular}
}
\vspace{-15pt}
\end{table}

Regarding the weighting in Equation~\ref{qua:feat-weight} in Sec.~\ref{sec:feat-weight}, we empirically selected the weights based on an ablation study conducted on both the Replica and ScanNet datasets. As shown in Table~\ref{tab:feat-weight-ablation}, assigning weights of 0.7 and 0.3 to $\mathbf{f}_{\text{image}}$ and $\mathbf{f}_{\text{text}}$, respectively, yields the best or near-best performance across all metrics.

\subsection{Detailed Object Classification in Abstraction}
we predefine three lists to support anchor object identification and map abstraction in Sec.~\ref{sec:map-abs}:

\begin{itemize}
\item \textbf{Volatile object examples:} A representative list of 12 categories selected from the GPT generated class list, including {[``backpack'', ``box'', ``clothes'', $\ldots$, ``indoor-plant'']}.

\item \textbf{Anchor object examples:} Similarly, we define a list of 12 anchor categories, such as {[``stool'', ``cabinet'', ``couch'', $\ldots$, ``bathroom-vanity'']}.

\item \textbf{Descriptive phrases for anchors:} We use manually defined phrases to describe anchor objects: {[``furniture that is not often moved'', ``furniture that is used for sitting'', ``furniture that is used for placing things'']}.

\end{itemize}

\begin{figure}[t!]
    \centering
    \includegraphics[width=1\linewidth]{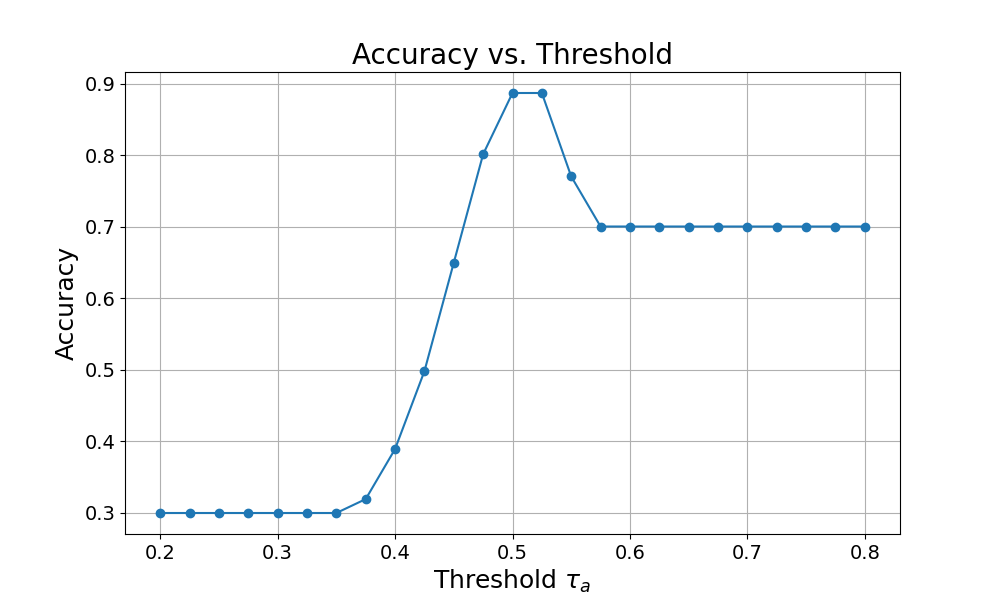}
    \caption{Binary classification accuracy of anchor vs. volatile objects w.r.t. threshold $\tau_a$ across Replica dataset}
    \label{fig:app-tau}
    \vspace{-15pt}
\end{figure}

All the objects from concrete map are embedded into a shared visual-semantic space using CLIP features, allowing us to classify them based on semantic feature similarity. To distinguish between anchor and volatile objects, we further use a two-step heuristic:

\begin{itemize}
    \item First, we compare each object's CLIP embedding with two representative category example lists—one for volatile objects and one for anchor objects. If the maximum similarity to one list exceeds the other by a margin $\Delta\tau = 0.05$, the object is assigned to the corresponding category.
    
    \item Second, for ambiguous cases where the margin condition is not met, we compute the similarity between the object and a list of descriptive phrases for anchors. If the similarity exceeds a threshold $\tau_a$, the object is classified as an anchor.
\end{itemize}

To determine the optimal threshold $\tau_a$, we conduct a validation experiment.
Specifically, we build concrete maps for 8 scenes in the Replica dataset and manually annotate each reconstructed object as either “anchor” or “volatile” to serve as ground truth.  
To evaluate the classification, we compute accuracy using the standard binary classification formula:
$\text{Acc} = \frac{TP + TN}{TP + TN + FP + FN}$, 
where a prediction is considered $True$ if it aligns with the manual annotation.
We sweep the threshold $\tau_a$ over the range $[0.2, 0.8]$ and plot the resulting accuracy in Figure~\ref{fig:app-tau}. Based on this curve, we select $\tau_a = 0.5$ as it yields the highest accuracy and thus provides a reasonable trade-off.

\begin{table*}[t!]
\centering
\caption{Open-vocabulary 3D Semantic Segmentation and Efficiency}
\label{tab:clip}
\resizebox{0.98\textwidth}{!}{%
\begin{tabular}{cc cccccccc}
\toprule
\textbf{Dataset} & \textbf{Method} & \textbf{CLIP-Backbone} & \textbf{mIoU $\uparrow$} & \textbf{FmIoU $\uparrow$} & \textbf{mAcc $\uparrow$} & \textbf{ODR} & \textbf{Avg. Mem $\downarrow$} & \textbf{Peak Mem $\downarrow$} & \textbf{TPF (s) $\downarrow$} \\
\midrule
\multirow{6}{*}{Replica}
& \multirow{2}{*}{ConceptGraphs} & ViT-H/14 & 0.1483 & 0.3124 & 0.3521 & 2.31 & 10044.6 & 23243.0 & 5.111 \\
&  & Mobile-CLIP & 0.1501 & 0.3858 & 0.3559 & 2.02 & 7148.9 & 23551.9 & 4.188 \\
& \multirow{2}{*}{HOV-SG}      & ViT-H/14 & 0.2129 & 0.4188 & 0.3794 & 3.13 & 70252.9 & 163238.5 & 45.056 \\
&        & Mobile-CLIP & 0.2050 & 0.4846 & 0.3835 & 3.81 & 73368.0 & 158126.6 & 42.005 \\
& \multirow{2}{*}{DualMap(Ours)} & ViT-H/14 & 0.2323 & 0.4859 & 0.3832 & \textbf{0.974} & 3281.5 & 4688.9 & 0.458 \\
&  & Mobile-CLIP & \textbf{0.2538} & \textbf{0.5207} & \textbf{0.4024} & 0.967 & \textbf{3095.2} & \textbf{4564.0} & \textbf{0.276} \\
\midrule
\multirow{6}{*}{ScanNet}
& \multirow{2}{*}{ConceptGraphs} & ViT-H/14 & 0.0646 & 0.1896 & 0.2394 & 7.75 & 13790.2 & 27086.0 & 7.605 \\
&  & Mobile-CLIP &  0.0882 & 0.3077 & 0.3538 & 6.97 & 9780.3 & 26155.2 & 6.301 \\
& \multirow{2}{*}{HOV-SG}        & ViT-H/14 & 0.1229 & 0.3105 & 0.3104 & 18.96 & 14561.2 & 44437.2 & 9.959 \\
&        & Mobile-CLIP & 0.1333 & \textbf{0.3381} & 0.3714 & 20.34 & 9223.0 & 25735.0 & 8.039 \\
& \multirow{2}{*}{DualMap(Ours)} & ViT-H/14 & \textbf{0.1611} & 0.3179 & 0.3632 & \textbf{2.55} & 2607.5 & 4428.6 & 0.306 \\
&  & Mobile-CLIP & 0.1604 & 0.3288 & \textbf{0.3794} & 2.56 & \textbf{2120.9} & \textbf{2820.2} & \textbf{0.163} \\
\bottomrule
\end{tabular}%
}
\end{table*}

\section{More Semantic Mapping Experiments}

\subsection{Results with Other CLIP Backbone}
\label{supp-vith}

The results presented in Table~\ref{tab:clip} highlight the advantages of using the Mobile-CLIP backbone for semantic segmentation tasks. Notably, methods employing Mobile-CLIP demonstrate superior accuracy and faster performance compared to those utilizing ViT-H/14 backbone. Our proposed method not only exhibits robustness across various CLIP backbones but also consistently outperforms other methods. Additionally, substituting Mobile-CLIP for ViT-H/14 significantly reduces memory usage and time costs, making it more resource-efficient for semantic segmentation applications.

\subsection{Results under Moving Humans}

\begin{figure*}[t!]
    \centering
    \includegraphics[width=0.8\linewidth]{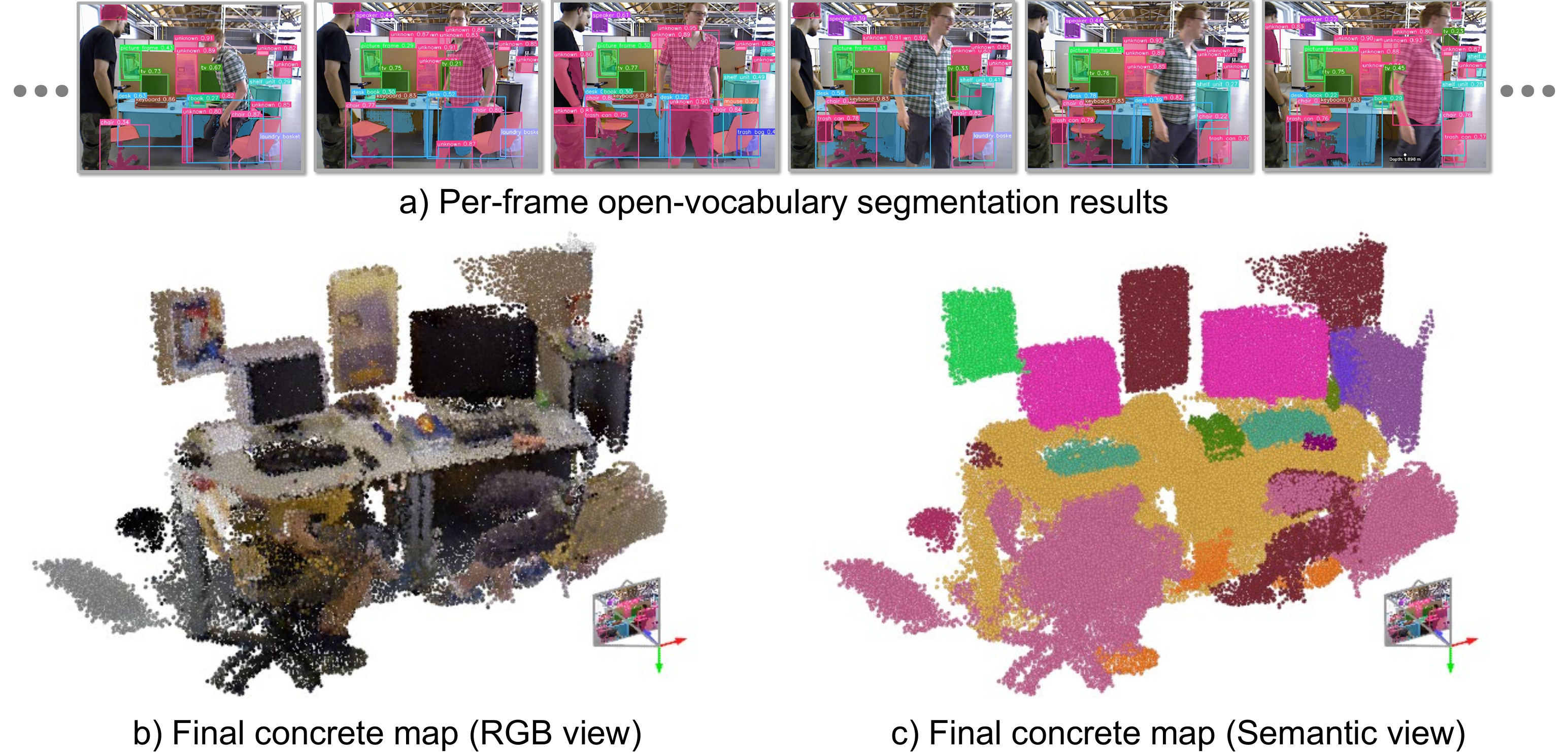}
    \caption{concrete mapping results on \texttt{freiburg3\_walking\_static}}
    \label{fig:tum-rgbd}
    \vspace{-10pt}
\end{figure*}

Regarding dynamic entities like humans, \textbf{DualMap is capable of generating a complete 3D semantic map without being affected by their movement}.
We conduct additional experiments on the\textit{ Dynamic Objects} split of the TUM RGBD dataset, widely used in the SLAM field. The qualitative results on \texttt{freiburg3\_walking\_static} sequence are shown in Figure~\ref{fig:tum-rgbd}. 
As illustrated in Figures~\ref{fig:tum-rgbd}-b and~\ref{fig:tum-rgbd}-c, the mapping results are unaffected by the movements of humans.
This robustness is attributed to our hybrid open-vocabulary segmentation pipeline and the object stability check mechanism (detailed in Sec.~III-A-1 and Sec.~III-B-3).
Figure~\ref{fig:tum-rgbd}-a shows part of the segmentation results, where humans, absent from YOLO’s class list, are occasionally segmented by FastSAM.
Since human movement leads to unstable segmentation, their related objects in concrete map are easily filtered out by the stability check due to the limited observations.

\begin{figure}[t]
    \centering
    \includegraphics[width=1\linewidth]{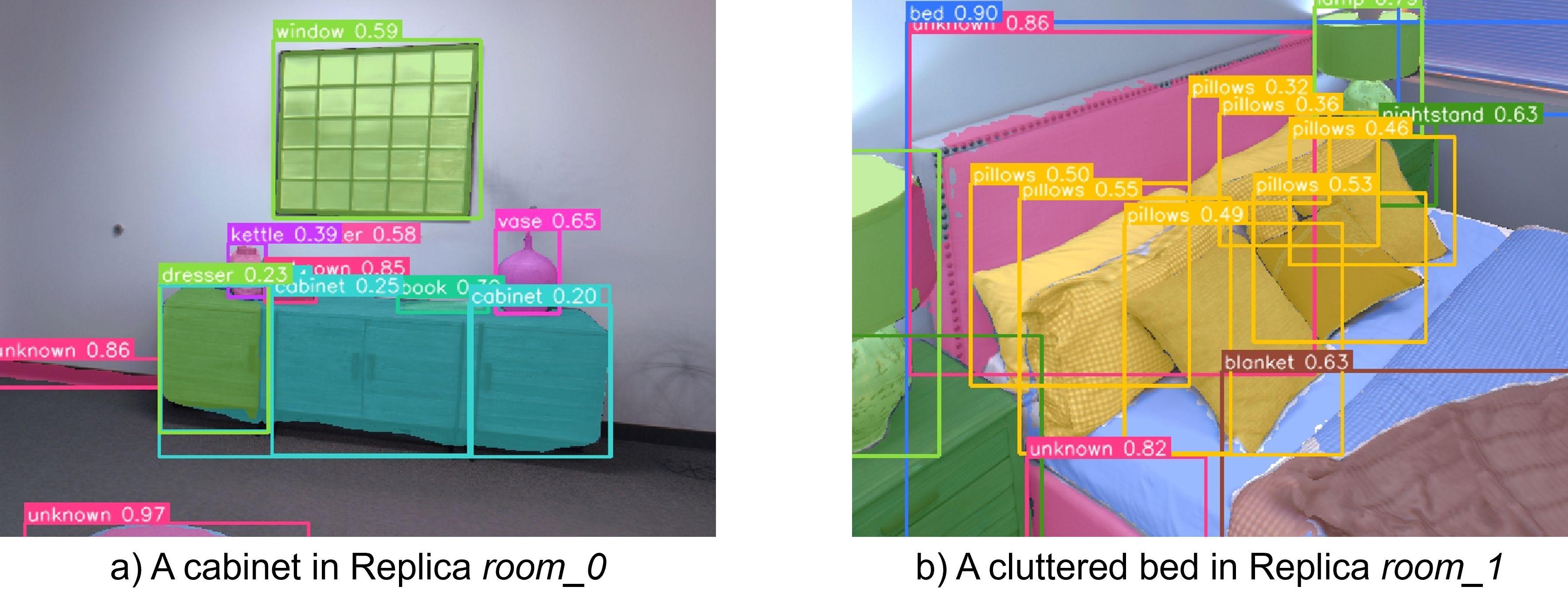}
    \caption{YOLO and FastSAM detections on two  frames.}
    \label{fig:yolo-output}
\end{figure}

\begin{figure}[t]
    \centering
    \includegraphics[width=1\linewidth]{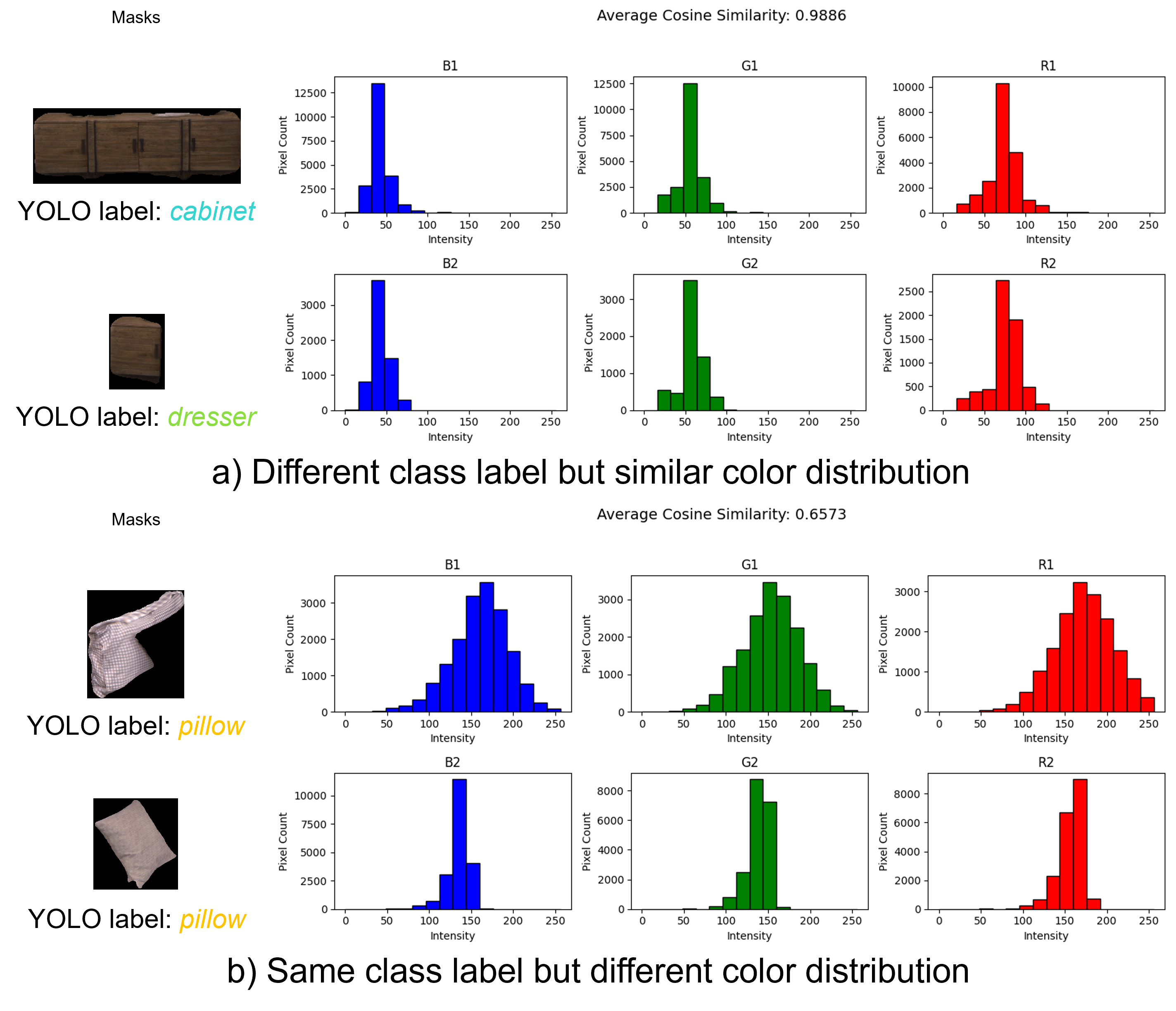}
    \caption{Visualization of BGR channel-wise color distributions for two overlapping segment pairs from \texttt{Replica}}
    \label{fig:color-dist}
\end{figure}

\subsection{Detection Refinement Details}
There are two factors that require refinement to improve the YOLO output. First, YOLO may assign \textbf{inconsistent class labels} to different parts of the same object, as shown in Figure~\ref{fig:yolo-output}-a, where a single cabinet is partially labeled as ``cabinet'' and partially as ``dresser.'' Label-based merging would fail to recognize them as one object. 
Second, in \textbf{cluttered scenes} such as Figure~\ref{fig:yolo-output}-b, multiple overlapping objects (e.g., pillows) often share the same class label, causing label-based methods to incorrectly merge distinct objects into one.

To address these issues, we adopt a color-based merging strategy. We first identify overlapping bounding box pairs. For each pair, we extract the pixel intensity histograms from the B, G, and R channels, dividing each into 16 equal-width bins: $[0, 16), [16, 32), \ldots, [240, 256)$. We then compute the cosine similarity between the corresponding channel histograms (e.g., $\mathbf{v}_{b1}$ and $\mathbf{v}_{b2}$ for blue), and take the average of the three channel-wise similarities as the overall color similarity.
If the average similarity exceeds a threshold (0.95 in our case), we treat the segments as visually consistent and merge them. This method successfully merges parts of the same object (Figure~\ref{fig:color-dist}-a) while avoiding incorrect merges in cluttered scenes (Figure~\ref{fig:color-dist}-b).

\section{More Performance Experiments}

\subsection{Runtime Decomposition}
\label{supp-runtime}

\begin{figure}[t!]
    \centering
    \begin{tikzpicture}
        \tikzstyle{every node}=[font=\small]
        \pie[
            text=legend,
            radius=2,
            color={gray!30, orange!20, blue!20}
        ]{
            57.9/Model Inference,
            28.7/Other Modules,
            13.5/Visualization
        }
    \end{tikzpicture}
    \vspace{20pt}
    \caption{System Runtime Breakdown on Replica}
    \label{fig:vis-a}
\end{figure}
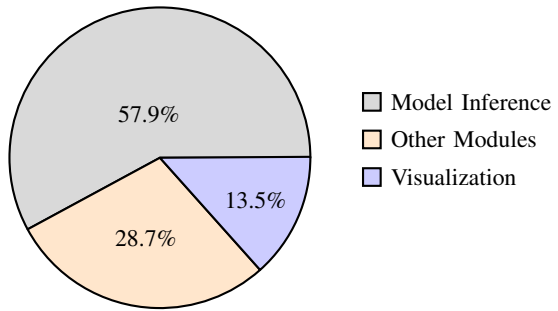

We break down the module-wise time cost on RTX 4090 Desktop in Table~\ref{tab:system-time} and Figure~\ref{fig:vis-a}. The results show that over \textbf{ 57\%} of the time per frame is spent on model inference (highlighted in grey). This indicates that applying inference optimizations such as TensorRT acceleration or model quantization can significantly reduce runtime with minimal accuracy impact.

\subsection{Experiments on Laptop and Different Resolutions}

While our main experiments were conducted on an RTX 4090 GPU Desktop, we conducted additional experiments on a \textbf{laptop} equipped with an RTX 3080 Laptop GPU. The additional system evaluation on Replica dataset is shown in Table~\ref{tab:gpu}. The results show that, at the same input resolution (1200×680), the RTX 3080 Laptop achieves comparable accuracy with a ~1.6× increase in latency (around 0.4s/frame), which remains acceptable for online applications. Furthermore, by reducing the input resolution to 640×360, the system achieves a good trade-off between accuracy and efficiency, with only a minor drop in accuracy (e.g., \textasciitilde3\% FmIoU loss) and comparable runtime performance to the results on the RTX 4090 Desktop. This suggests that appropriately adjusting input resolution offers a practical path for deployment on lower-cost devices.

\begin{table}[t!]
\centering
\caption{Time Decomposition on Replica with 4090}
\label{tab:system-time}
\vspace{0.5em}
\resizebox{\columnwidth}{!}{%
\begin{tabular}{c|c|c|c}
\toprule
\textbf{TPF (s)} & \textbf{Modules} & \textbf{Submodules} & \textbf{Time (s)} \\
\midrule
\multirow{10}{*}{0.2524} 
& \multirow{6}{*}{\shortstack{Observation\\ Generation$^*$ \\ 0.2183}} 
& \cellcolor{gray!20}FastSAM $\dagger$ & \cellcolor{gray!20}0.0592 \\
&& \cellcolor{gray!20}YOLO+MobileSAM $\dagger$ & \cellcolor{gray!20}0.0373 \\
&& Detection Filter & 0.0372 \\
&& Create Obj. Pointcloud$\ddagger$ & 0.0866 \\
&& \cellcolor{gray!20}CLIP$\ddagger$ & \cellcolor{gray!20}0.0619 \\
& & Observation Formatting & 0.0187 \\
\cmidrule(lr){2-4}
& \multirow{2}{*}{\shortstack{Mapping$^*$ \\ 0.0485}}  & Concrete Mapping & 0.0485 \\
& & Abstraction & 0.0000 \\
\cmidrule(lr){2-4}
& \shortstack{Visualization \\ 0.0341} & -- & 0.0341 \\
\bottomrule
\end{tabular}
}
\vspace{10pt}
\parbox{\linewidth}{\centering\footnotesize
$^{*}$, $^{\dagger}$, and $^{\ddagger}$ indicate parallel execution.}
\end{table}

\begin{table*}[t!]
\centering
\caption{Accuracy and Efficiency Across GPU and Resolution Settings on Replica}
\label{tab:gpu}
\begin{tabular}{c c c c c c c c}
\toprule
\textbf{GPU} & \textbf{Resolution} & \textbf{FmIoU $\uparrow$} & \textbf{mAcc $\uparrow$} & \textbf{mIoU $\uparrow$} & \textbf{TPF (s) $\downarrow$} & \textbf{Rel. FmIoU$^\ast$ $\uparrow$} & \textbf{Rel. TPF$^\ast$ $\downarrow$} \\
\midrule
RTX 4090 Desktop & 1200×680 & 0.5508 & 0.4251 & 0.2508 & 0.2524 & 100.00\% & 100.00\% \\
\midrule
\multirow{4}{*}{RTX 3080 Laptop} 
& 1200×680 & 0.5503 & 0.4256 & 0.2502 & 0.4045 & 99.92\%  & 160.26\% \\
& 960×540  & 0.5507 & 0.4259 & 0.2526 & 0.3221 & 99.98\%  & 127.61\% \\
& \cellcolor{gray!20}640×360  & \cellcolor{gray!20}0.5341 & \cellcolor{gray!20}0.3941 & \cellcolor{gray!20}0.2428 & \cellcolor{gray!20}0.2717 & \cellcolor{gray!20}96.97\%  & \cellcolor{gray!20}107.65\% \\
& 320×180  & 0.2646 & 0.1437 & 0.0801 & 0.2538 & 48.04\%  & 100.55\% \\
\bottomrule
\end{tabular}

\vspace{4pt}
\footnotesize{\textsuperscript{$\ast$} Relative FmIOU and Relative TPF (Time Per Frame) are computed with respect to the 4090 Desktop at 1200$\times$680 resolution.}
\end{table*}

\begin{figure*}[t!]
    \centering
    \includegraphics[width=1\linewidth]{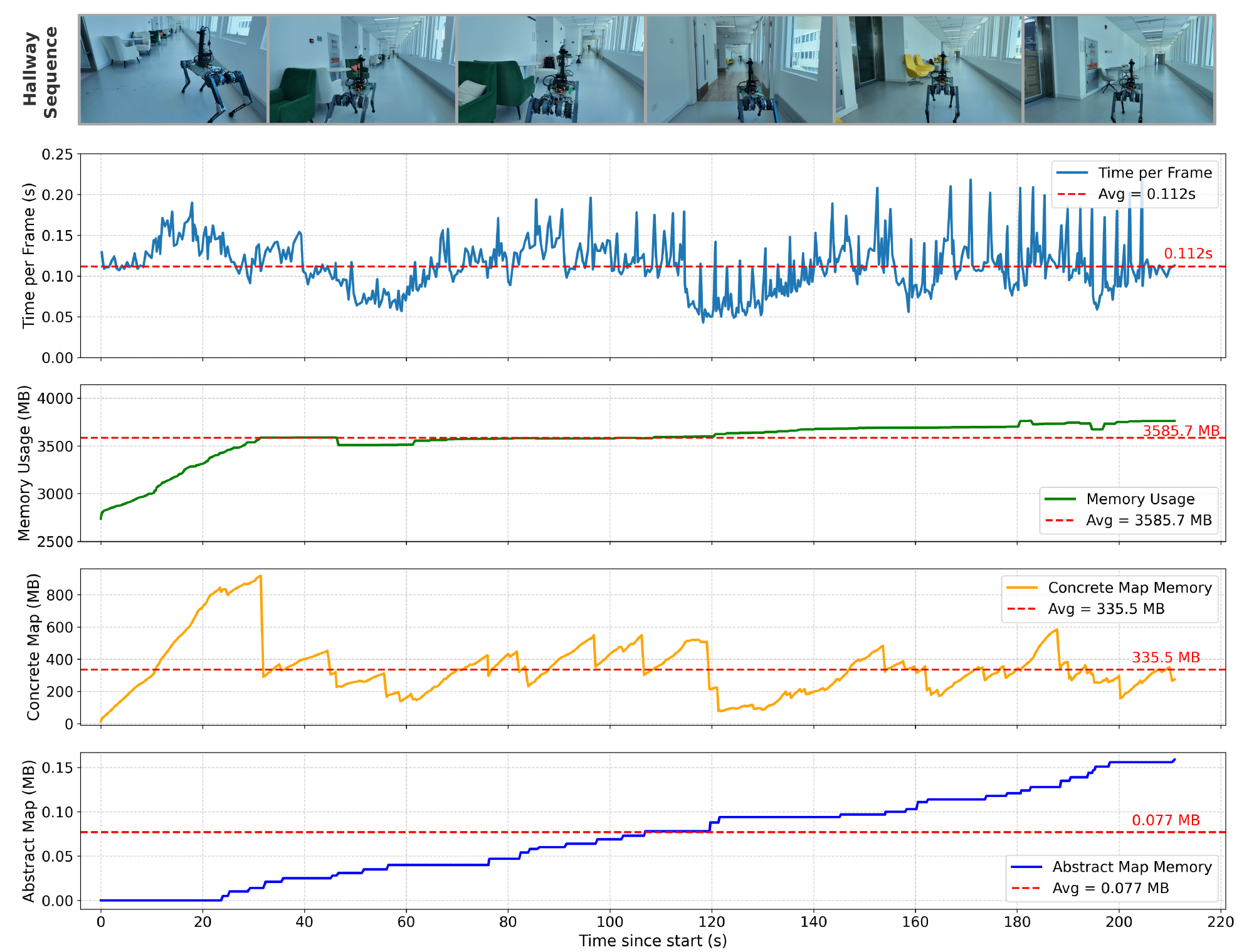}
    \caption{Memory usage of concrete and abstract maps in a\textbf{ hallway sequence} (212s).}
    \label{fig:app-hallway-con-abs}
\end{figure*}

\subsection{Resources Usage in Long Operation}

\begin{figure*}[t!]
    \centering
    \includegraphics[width=1\linewidth]{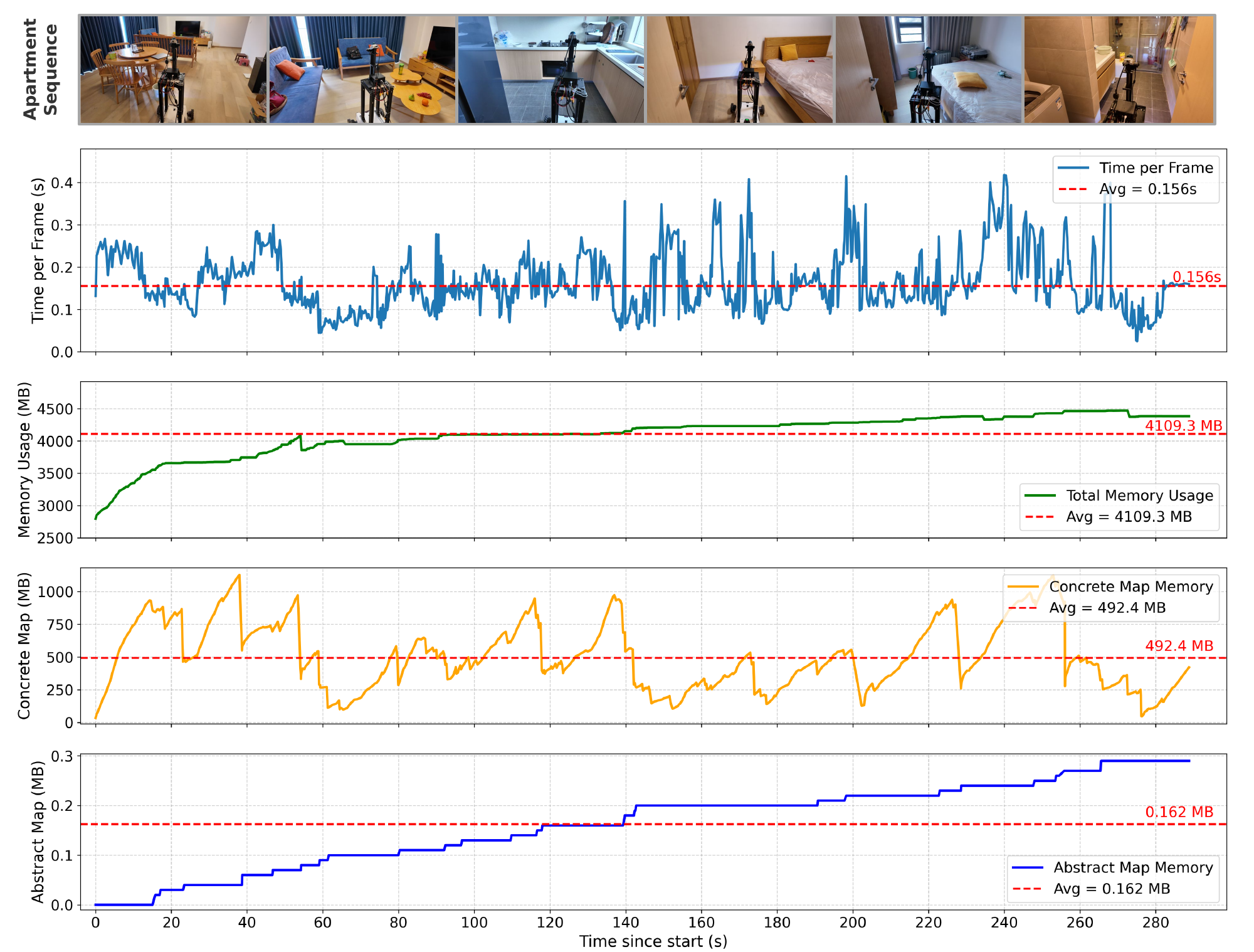}
    \caption{Memory usage of concrete and abstract maps in an\textbf{ apartment sequence} (286s).}
    \label{fig:app-apartment-con-abs}
\end{figure*}

We additionally conducted long-term mapping experiments in both a structured hallway and a cluttered apartment. The experiments were performed using an RTX 4090 Desktop GPU with an input resolution of 1280×720. As shown in top two rows in Figure~\ref{fig:app-hallway-con-abs} and Figure~\ref{fig:app-apartment-con-abs}, the average runtime per frame remains stable (\textbf{0.112s} and\textbf{ 0.156s}, respectively), and memory usage (CPU RAM) grows slowly and stays bounded (\textbf{around 3500–4200MB}), even in more complex environments. 
This is benefit from our dual-map design, the concrete map memory size remains stable, resulting in consistent time cost in matching. Although the abstract map expands, its memory usage is minimal (only a few hundred KB). While the dual map maintains stable memory usage, the overall system memory gradually increases due to the cost of visualization. The detailed memory usage of concrete map and abstract map can be found in bottom two rows in Figure~\ref{fig:app-hallway-con-abs} and Figure~\ref{fig:app-apartment-con-abs}.
The testing results demonstrate that \textbf{DualMap scales well over time without ballooning in runtime or memory usage}.

\section{Acknowledgment}
We would like to express our sincere gratitude to Pengxu Hou and Runze Yu for generously allowing us to use their experimental platform for the initial system validation. Special thanks to Guowei Huai, Qingyun Wang, Yingxi Lin, and Boyu Zhou for their valuable assistance during the real-world experiments. We also appreciate the insightful suggestions and feedback provided by Bonan Liu, Shibo Wang, Zhengmao He, and Handi Yin, which greatly contributed to this project.

\end{document}